\documentclass[10pt,onecolumn]{article}
\usepackage[margin=1in]{geometry}
\usepackage{amsmath,amssymb,amsfonts}
\usepackage{booktabs}
\usepackage{multirow}
\usepackage{graphicx}
\usepackage{xcolor}
\usepackage{hyperref}
\usepackage{adjustbox}
\usepackage{array}
\usepackage{times}
\usepackage{microtype}
\usepackage{enumitem}
\usepackage{algorithm}
\usepackage{algpseudocode}
\usepackage{caption}
\usepackage{subcaption}
\usepackage{tabularx}
\usepackage[numbers,compress]{natbib}

\definecolor{myblue}{HTML}{1F4E79}
\definecolor{myteal}{HTML}{1D6A72}
\definecolor{mygray}{HTML}{595959}

\hypersetup{colorlinks=true,linkcolor=myblue,citecolor=myteal,urlcolor=myteal}

\usepackage{titlesec}
\titleformat{\section}{\large\bfseries\color{myblue}}{{\thesection}}{0.5em}{}
\titleformat{\subsection}{\normalsize\bfseries\color{myteal}}{{\thesubsection}}{0.5em}{}
\titleformat{\subsubsection}{\normalsize\bfseries}{{\thesubsubsection}}{0.5em}{}

\setlist[itemize]{noitemsep,topsep=2pt,leftmargin=*}
\setlist[enumerate]{noitemsep,topsep=2pt,leftmargin=*}

\newcommand{\R}{\mathbb{R}}

\title{\textbf{OmicSync: Reliability-Aware Spatial Multi-Omics Clustering with
Evidence-Constrained LLM Reasoning}}

\author{
Rabeya Tus Sadia\textsuperscript{1}\quad
Qiang Ye\textsuperscript{2}\quad
Qiang Cheng\textsuperscript{1,*}\\[4pt]
\textsuperscript{1}Department of Computer Science, University of Kentucky\\
\textsuperscript{2}Department of Mathematics, University of Kentucky\\[2pt]
\textsuperscript{*}Corresponding author: 
\texttt{qiang.cheng@uky.edu}\\[2pt]
}

\date{}

\begin{document}
\maketitle

\begin{abstract}
Spatial multi-omics technologies can jointly profile gene expression, surface proteins, and histology at each tissue spot, but most spatial domain discovery methods return only a partition of spots into domains, without indicating which assignments are reliable, which modality drove each decision, or why a particular domain assignment should be trusted.
We present \textbf{OmicSync}, a reliability-aware spatial multi-omics framework that couples unsupervised domain clustering with evidence-constrained LLM reasoning through model-derived per-spot signals, including soft assignment confidence, epistemic routing uncertainty, and modality-routing weights. These signals are converted into structured evidence dictionaries and passed to a five-strategy reasoning module that provides standard, stepwise, counterfactual, contrastive, and uncertainty-focused explanations. OmicSync combines a KAN-GCN backbone with spatial encoding, cross-modal fusion, uncertainty-aware routing, cell-type supervision, and missing-modality imputation.
We further introduce \textbf{OmicSync-R}, a variant that closes the reasoning--clustering loop by using automatically computed reasoning-quality scores as REINFORCE rewards during training. This formulation allows reasoning coherence to shape the learned latent structure without requiring gradients to propagate through the language model.
Across four 10x CytAssist FFPE spatial proteomics benchmarks, OmicSync achieves the best average rank on Human Tonsil (1.44), Glioblastoma (1.78), and Tonsil Add-on (1.22), and the second-best average rank on Human Breast Cancer (2.33). OmicSync-R further improves the ARI on Human Breast Cancer from 45.73 to 46.72 and outperforms existing methods on six of the nine clustering metrics. Together, OmicSync and OmicSync-R move spatial domain discovery beyond opaque partitioning toward tissue-domain analysis that is reliability-aware, auditable at the spot level, and guided by evidence-constrained reasoning.
\end{abstract}

\section{Introduction}
\label{sec:intro}

Spatial multi-omics technologies now enable the joint profiling of gene expression, surface protein abundance, and tissue morphology at spatially resolved tissue spots, preserving the spatial context lost in dissociative single-cell protocols~\citep{staahl2016visualization, lundberg2019spatial}. Recent platforms such as 10x Genomics CytAssist FFPE extend this capability to formalin-fixed, paraffin-embedded clinical tissues, generating multimodal datasets comprising thousands of spots, each characterized by tens of thousands of genes, dozens of antibody-derived tag (ADT) protein markers, and a high-resolution histology image patch from the same physical location.

Unsupervised spatial domain discovery, which partitions spots into biologically coherent tissue domains, is a foundational analysis step for downstream tasks such as differential expression analysis, ligand--receptor inference, and cell-state mapping~\citep{zhao2021spatial, xu2022unsupervised}. A growing family of methods has been developed for this task: GROVER~\citep{grover2024} uses KAN-based graph convolutions with dual spatial and feature adjacency matrices; MISO~\citep{miso2024} performs multi-scale integration; SpatialGlue~\citep{spatialglue2024} fuses modalities through cross-attention; and COSMOS~\citep{cosmos2024} applies contrastive multi-view learning. 
Despite these methodological advances, existing approaches generally return only one type of output, namely, a flat partition of spots into domains, without indicating \emph{which} assignments are reliable, \emph{why} a spot was assigned to a particular domain, or \emph{which} molecular modality primarily drove each assignment.

This limitation is consequential because downstream analyses that treat all spot assignments as equally reliable may propagate clustering errors into biological conclusions, particularly for boundary spots where adjacent domains overlap or for spots affected by degraded data quality. Quantifying \emph{how confident} the model is in each assignment, and explaining \emph{why}, is therefore not merely desirable but crucial for the responsible interpretation and use of spatial clustering outputs in biomedical research.

\paragraph{Overview.}
We present \textbf{OmicSync}, a reliability-aware spatial multi-omics
framework that links tissue-domain discovery with evidence-constrained
natural-language auditing.
Rather than treating interpretability as an external visualization step,
OmicSync extracts three core model-derived reliability signals for each
spot:
(i) soft assignment confidence from a Student's $t$-distribution-based
clustering head;
(ii) epistemic routing uncertainty estimated through MC dropout in the
mixture-of-experts gate; and
(iii) modality-routing weights that indicate the relative weighting of
RNA, ADT, and histology in the fused representation.
These signals, together with feature-level and spatial-neighbourhood
evidence, are assembled into a structured evidence dictionary and used
by the reasoning module to generate evidence-constrained, per-spot
reliability reports.
We further propose \textbf{OmicSync-R}, a variant that closes the
reasoning--clustering loop by using automatically computed
reasoning-quality scores as REINFORCE rewards during training.
This formulation allows reasoning coherence to shape the learned latent
structure without requiring gradients to propagate through the language
model.

\paragraph{Contributions.}
Our main contributions are as follows:
\begin{itemize}
\item A reliability-aware spatial multi-omics clustering framework,
OmicSync, built on a KAN-GCN backbone. Because spatial multi-omics data
combine heterogeneous modalities with missing entries and complementary
biological signals, robust domain discovery requires joint alignment,
uncertainty-aware fusion, and tolerance to incomplete modality profiles.
OmicSync addresses these needs by integrating spatial position encoding,
per-modality graph encoding, cross-modal Transformer fusion,
uncertainty-aware mixture-of-experts routing, cluster-assignment prediction,
semi-supervised cell-type regularisation, and missing-modality imputation in
a unified architecture.

\item An adaptive spatial training strategy that combines an adaptive
spatial exclusion radius with a decaying contrastive-loss schedule to
mitigate the trade-off between adjusted Rand index and silhouette
coefficient across heterogeneous and homogeneous tissue datasets.

\item An evidence-constrained, five-strategy reasoning module that converts
model-derived evidence into per-spot reliability reports. The supplied
evidence includes assignment confidence, modality-routing weights, marker
evidence, uncertainty estimates, and neighbourhood composition, and the
reasoning strategies include standard, stepwise, counterfactual,
contrastive, and uncertainty-focused reasoning.

\item OmicSync-R, a REINFORCE-based reasoning-guided extension that feeds
automatically computed reasoning-quality scores back into model training,
allowing evidence-grounded reasoning coherence to influence the learned
latent structure without differentiating through the language model.

\item A quantitative clustering and reliability evaluation on four 10x
CytAssist FFPE benchmarks. OmicSync achieves the best average rank on
three datasets and the second-best average rank on Human Breast Cancer.
The reliability audit further shows that high-confidence spots exhibit
lower epistemic routing uncertainty and greater spatial-neighbourhood
homogeneity than low-confidence spots.

\item A proof-of-concept evaluation of reasoning-guided training on Human
Breast Cancer, the only benchmark where base OmicSync does not achieve the
best overall rank. OmicSync-R improves ARI from 45.73 to 46.72 and
outperforms GROVER, the most competitive non-OmicSync baseline, on six of
the nine clustering metrics evaluated on that dataset.
\end{itemize}

\section{Background and Related Work}
\label{sec:related}

\subsection{Spatial multi-omics domain discovery}

Early approaches to spatial domain discovery primarily analysed RNA
measurements and combined dimensionality reduction with conventional
clustering methods such as $k$-means or Leiden clustering
~\citep{maynard2021transcriptome}.
Graph-based methods such as STAGATE~\citep{dong2022deciphering}
and GraphST~\citep{long2023spatially} incorporate spatial proximity
through graph neural networks, thereby improving the spatial coherence
of the resulting domains.
More recent methods integrate multiple molecular and imaging modalities.
SpatialGlue~\citep{spatialglue2024} and MISO~\citep{miso2024}
integrate RNA and protein measurements, whereas
GROVER~\citep{grover2024} and COSMOS~\citep{cosmos2024}
jointly model RNA, protein, and histology to improve multimodal
representation learning and tissue-domain identification.
OmicSync employs KAN-based graph convolutions over spatial and feature adjacency matrices, followed by attention-based multimodal fusion to integrate complementary information across modalities. Building on this architecture, OmicSync further incorporates reliability estimation, uncertainty-aware routing, evidence-constrained reasoning, and reasoning-guided model refinement to improve both the robustness and interpretability of spatial multi-omics analysis.

\subsection{Explainability in spatial omics}

Explainability in spatial omics remains relatively underdeveloped.
Attention-weight visualisation and SHAP-based feature attribution have
been used to identify influential features in biological prediction
models~\citep{lundberg2020local}.
Although such methods can reveal which inputs influence a prediction,
they generally do not provide spot-specific explanations of clustering
assignments grounded in multiple internal reliability signals.
Large language models (LLMs) have also been applied in single-cell biology~\citep{ge2025deep} for
tasks such as cell-type annotation~\citep{wang2023gpt4} and pathway
summarisation~\citep{lievin2022can}.
To our knowledge, however, existing work has not coupled an LLM with a
trained spatial clustering model through structured, model-derived
evidence to generate reliability-aware explanations for individual
spots.
OmicSync addresses this gap by constraining natural-language reasoning
with confidence, uncertainty, modality-routing, feature-level, and
spatial-neighbourhood evidence extracted from the clustering model.

\subsection{Mixture-of-experts routing and uncertainty estimation}

Sparse mixture-of-experts routing~\citep{shazeer2017outrageously}
has been used in multimodal learning to route inputs among
modality-specialised experts.
MC dropout~\citep{gal2016dropout} provides a computationally tractable
approximation to Bayesian inference and enables epistemic uncertainty
to be estimated by repeatedly evaluating a dropout-equipped network.
OmicSync combines these ideas by evaluating its mixture-of-experts
gating network multiple times with dropout enabled.
The mean gate outputs provide per-spot modality-routing weights, while
their variation across stochastic forward passes quantifies epistemic
uncertainty in the routing decision.
This design therefore provides both an interpretable estimate of
modality reliance and a measure of the stability of that estimate.

\subsection{Reinforcement learning with non-differentiable feedback}

REINFORCE~\citep{williams1992simple} is a policy-gradient estimator
that enables non-differentiable reward signals to influence neural
network training.
In natural language processing, policy-gradient methods have been used
to optimise sequence-level objectives that cannot be directly
differentiated through token-generation decisions
~\citep{ranzato2016sequence}.
OmicSync-R adapts this principle by treating automatically computed
reasoning-quality scores as rewards and feeding them back into the
clustering model during training.
This mechanism allows reasoning coherence to shape the learned latent
structure without propagating gradients through the language model.
To our knowledge, reasoning-quality feedback has not previously been
used in this manner to refine spatial multi-omics domain clustering.

\section{Problem Formulation}
\label{sec:problem}

Consider a spatial multi-omics sample containing $N$ tissue spots.
Each spot is characterized by three modality-specific feature
representations after appropriate preprocessing:
$\mathbf{X}_1 \in \mathbb{R}^{N \times d_1}$ for RNA features,
$\mathbf{X}_2 \in \mathbb{R}^{N \times d_2}$ for ADT protein features,
and $\mathbf{X}_3 \in \mathbb{R}^{N \times d_3}$ for histology features.
The corresponding spatial coordinates are denoted by
$\mathbf{C} \in \mathbb{R}^{N \times 2}$.
For each modality $m \in \{1,2,3\}$, we construct a spatial adjacency
matrix
$\mathbf{A}^{s}_m \in \mathbb{R}_{\geq 0}^{N \times N}$,
which encodes physical proximity between spots, and a feature adjacency
matrix
$\mathbf{A}^{f}_m \in \mathbb{R}_{\geq 0}^{N \times N}$,
which encodes similarity in the corresponding modality-specific
feature space.

\textbf{Task~A (Spatial domain clustering and reliability estimation).}
Given the multimodal inputs
\begin{equation}
\left\{
\mathbf{X}_m,
\mathbf{A}^{s}_m,
\mathbf{A}^{f}_m
\right\}_{m=1}^{3}
\quad \text{and} \quad
\mathbf{C},
\end{equation}
the objective is to learn a $d$-dimensional shared latent representation
$\mathbf{Z} \in \mathbb{R}^{N \times d}$ and a soft assignment matrix
$\mathbf{Q} \in [0,1]^{N \times K}$ that partitions the $N$ spots into
$K$ spatial domains.
Each row of $\mathbf{Q}$ represents a probability distribution over
the $K$ domains and therefore satisfies
\begin{equation}
\sum_{k=1}^{K} q_{ik} = 1,
\qquad i=1,\ldots,N,
\end{equation}
where $q_{ik}$ denotes the $(i,k)$-th entry of $\mathbf{Q}$.
The hard domain assignment and assignment confidence for spot $i$ are
defined as
\begin{equation}
\hat{k}_i = \arg\max_{1 \leq k \leq K} q_{ik},
\qquad
c_i = \max_{1 \leq k \leq K} q_{ik}.
\end{equation}
Collecting the confidence scores for all spots gives
$\mathbf{c} \in [0,1]^N$.

In addition, the model estimates an epistemic routing uncertainty vector
$\mathbf{u} \in \mathbb{R}_{\geq 0}^{N}$ and a modality-routing matrix
$\mathbf{G} \in [0,1]^{N \times 3}$.
The entry $G_{im}$ denotes the routing weight assigned to modality $m$
for spot $i$, where
\begin{equation}
\sum_{m=1}^{3} G_{im} = 1,
\qquad i=1,\ldots,N.
\end{equation}
Thus, each spot is associated with a domain assignment and three core
reliability signals: soft assignment confidence, epistemic routing
uncertainty, and modality-routing weights.

\textbf{Task~B (Optional annotation and auxiliary regularisation).}
When spot-level cell-type annotations are available, let
$\mathbf{Y} \in \{0,1\}^{N \times L}$ denote the corresponding label
matrix for $L$ cell types.
The objective is to learn an annotation function
\begin{equation}
f_{\mathrm{ann}}:
\mathbb{R}^{d}
\rightarrow
[0,1]^{L},
\end{equation}
which predicts a cell-type distribution from the shared latent
representation of each spot.

Because the datasets considered in this study do not provide verified
spot-level cell-type labels, Task~B is instantiated using pseudo-labels
derived from Leiden community detection on the RNA representation.
We use a subset of these Leiden-derived labels as auxiliary
classification targets for the annotation head.
Under this setting, Task~B serves as a semi-supervised consistency
regularizer for the learned representation rather than as an
independently validated cell-type annotation task.

\textbf{Task~C (Evidence-constrained reasoning).}
For a queried subset of spots
$\mathcal{S} \subseteq \{1,\ldots,N\}$, a structured evidence
dictionary is constructed for each spot $i \in \mathcal{S}$ using the
outputs and intermediate quantities of Task~A.
The evidence includes assignment confidence, epistemic routing
uncertainty, modality-routing weights, feature-level evidence, and
spatial-neighbourhood evidence.

Let $\mathcal{R}$ denote the set of reasoning strategies.
For each queried spot $i \in \mathcal{S}$ and each reasoning strategy
$r \in \mathcal{R}$, the reasoning module generates a natural-language
reliability report $J_i^{(r)}$.
The complete set of reports is written as
\begin{equation}
\mathcal{J}
=
\{
J_i^{(r)}
\mid
i \in \mathcal{S},\;
r \in \mathcal{R}
\}.
\end{equation}
Each report is constrained by and supported by the corresponding
model-derived evidence.

\section{OmicSync Architecture}
\label{sec:method}

\begin{figure*}[t]
\centering
\includegraphics[width=\textwidth]{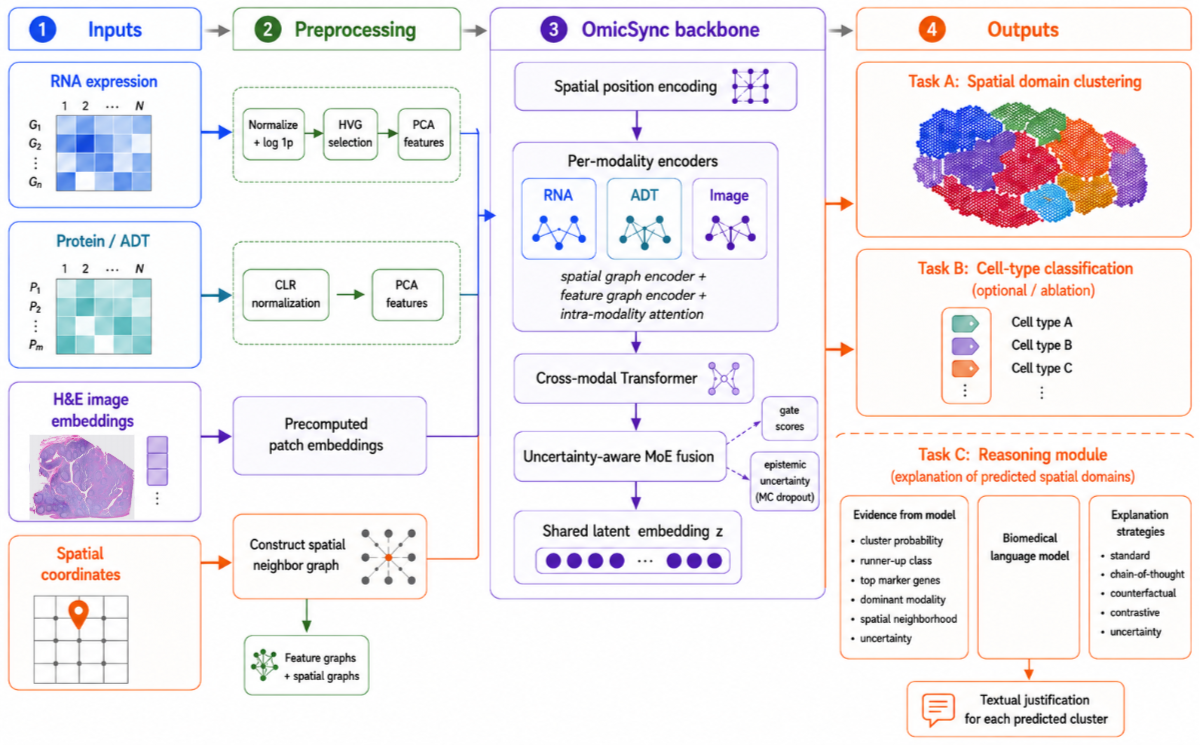}
\caption{
Overview of the OmicSync architecture and training workflow.
OmicSync integrates RNA expression, ADT protein abundance, H\&E image
embeddings, and spatial coordinates within a unified spatial multi-omics
framework.
RNA and ADT profiles are normalised and projected into PCA feature spaces,
histological information is represented using pretrained image-patch
embeddings, and spatial coordinates are used to construct neighbourhood
graphs.
The OmicSync backbone combines spatial position encoding,
modality-specific graph encoders, intra-modality attention,
cross-modal Transformer fusion, and uncertainty-aware
mixture-of-experts routing to obtain a shared latent representation.
The representation is optimised through reconstruction, clustering,
optional auxiliary annotation, missing-modality imputation, and
uncertainty-regularisation objectives.
The framework outputs spatial domain assignments and model-derived
per-spot reliability signals, which are assembled into structured
evidence dictionaries and passed to an evidence-constrained reasoning
module to generate natural-language reliability reports.
}
\label{fig:arch}
\end{figure*}

Figure~\ref{fig:arch} presents the overall architecture and training
workflow of OmicSync.
The framework first preprocesses RNA, ADT, histology, and spatial
coordinate information into modality-specific feature representations
and neighbourhood graphs.
These inputs are then processed by spatially informed graph encoders,
intra-modality attention, cross-modal Transformer fusion, and an
uncertainty-aware mixture-of-experts router to produce a shared latent
representation.
The learned representation supports spatial domain clustering,
optional auxiliary annotation, missing-modality imputation, and the
extraction of per-spot reliability evidence.
The resulting confidence, uncertainty, modality-routing, feature-level,
and spatial-neighbourhood signals are assembled into structured evidence
dictionaries for evidence-constrained natural-language reasoning.

\subsection{Preprocessing pipeline}

\subsubsection{RNA}
Raw counts are normalised to a library size of $10^4$ counts per spot
and subsequently transformed using $\log(1+x)$.
The top 3,000 highly variable genes are selected using the Seurat v3
method~\citep{stuart2019comprehensive}, with the Seurat flavour used
as a fallback when \texttt{scikit-misc} is unavailable.
The highly variable gene matrix is reduced to
$d_1 \leq 49$ principal components, forming $\mathbf{X}_1$.

\subsubsection{Protein abundance (ADT)}
Antibody-derived tag (ADT) counts undergo centered log-ratio (CLR)
normalisation to mitigate compositional effects in protein-abundance
measurements and are subsequently reduced to $d_2 \leq 49$ principal
components, forming $\mathbf{X}_2$.
NaN, infinite, and negative values are replaced with zero before
normalisation.

\subsubsection{Histology}
A $224 \times 224$ H\&E image patch centred at the pixel coordinate of
each spot is extracted from the high-resolution tissue image.
Patches extending beyond the image boundary are padded with white pixels.
Each patch is processed using the pathology foundation model UNI~\cite{chen2024uni}, a ViT-L/16 model pretrained on more than 100,000 pathology slides.
The resulting 1,024-dimensional CLS-token embedding is L2-normalised
to form $\mathbf{X}_3$.
Embeddings containing NaN values, which typically arise from
predominantly white boundary patches, are replaced with zero vectors.

\subsection{Component 1: SpatialPositionEncoder}

Absolute spot coordinates are encoded using a sinusoidal 2-D positional
encoding:
\begin{equation}
  \mathrm{PE}_m(\mathbf{c}_i) =
  W_m^{\mathrm{PE}}
  \bigl[
    \sin(c_{i,x}\mathbf{f}),\,
    \cos(c_{i,x}\mathbf{f}),\,
    \sin(c_{i,y}\mathbf{f}),\,
    \cos(c_{i,y}\mathbf{f})
  \bigr]^{\top}
  \in \R^{d_m},
\end{equation}
where $\mathbf{c}_i=(c_{i,x},c_{i,y})$ denotes the 2-D coordinate of
spot $i$, $\mathbf{f}\in\R^F$ is a fixed vector containing $F=32$
predefined log-spaced frequencies, and
$W_m^{\mathrm{PE}}\in\R^{d_m\times 4F}$ is a learned projection.

This positional bias is added to each modality's feature vector before graph
convolution, so that two spots with identical molecular profiles but different
tissue locations, such as the edge versus the center of a germinal center,
receive distinguishable representations. This design allows the model to
incorporate spatial context and tissue architecture that are not captured by
molecular composition alone.

\subsection{Component 2: KAN-GCN encoders and within-modality attention}

For each modality $m \in \{1,2,3\}$, a shared-weight KAN-GCN encoder is
applied over both the spatial adjacency matrix and the feature adjacency
matrix:
\begin{equation}
  \mathbf{e}_m^s = \mathrm{KAN}(A^s_m \mathbf{X}_m W_m^{\mathrm{GCN}}),\quad
  \mathbf{e}_m^f = \mathrm{KAN}(A^f_m \mathbf{X}_m W_m^{\mathrm{GCN}}).
\end{equation}
The two embeddings are then merged by a within-modality attention layer,
which adaptively weights spatial and feature-based evidence to produce
$\mathbf{e}_m \in \R^{N \times d}$.
This backbone follows the GROVER design and is kept unchanged, thereby
isolating the contribution of the five new components introduced in
OmicSync.

\subsection{Component 3: CrossModalTransformer}

The three per-modality embeddings are stacked into a length-3 token
sequence $\mathbf{T} \in \R^{N \times 3 \times d}$ and processed by
a multi-head self-attention Transformer encoder with learnable
modality-type embeddings:
\begin{equation}
  \mathbf{T}' =
    \mathrm{TransEnc}\bigl(\mathbf{T} + \mathbf{M}\bigr)
    + \mathbf{T},
  \qquad
  [\mathbf{e}'_1, \mathbf{e}'_2, \mathbf{e}'_3] = \mathbf{T}',
\end{equation}
where $\mathbf{M} \in \R^{N \times 3 \times d}$ is obtained by
broadcasting the learnable modality-type embeddings
$\mathrm{Emb}([0,1,2]) \in \R^{3 \times d}$ across all $N$ spots.
This cross-modal Transformer allows RNA, protein, and histology tokens
from the same spot to attend to one another \emph{before} fusion,
thereby enriching each modality with cross-modal context rather than
combining modalities only at the gating stage.
We use $n_\mathrm{heads}=4$ attention heads and $n_\mathrm{layers}=2$
Transformer layers.

\subsection{Component 4: UncertaintyMoE}
\label{sec:umoe}

The UncertaintyMoE fuses the three cross-modally enriched embeddings
through a learned gating network evaluated under MC-Dropout, a standard
approximate Bayesian technique that keeps dropout active at inference
time and uses multiple stochastic forward passes to estimate epistemic
uncertainty. We use $T=10$ stochastic samples. For spot $i$, we first compute the averaged cross-modal token
\begin{equation}
  \bar{\mathbf{e}}_i =
  \frac{1}{3}\left(\mathbf{e}'_{1,i}+\mathbf{e}'_{2,i}
  +\mathbf{e}'_{3,i}\right).
\end{equation}
For each MC-Dropout sample $t$, the gating network produces
\begin{equation}
  \mathbf{g}_{i,t} =
    \mathrm{softmax}\bigl(W_g \,\mathrm{Dropout}_t(\bar{\mathbf{e}}_i)\bigr)
    \in [0,1]^3,
\end{equation}
where $\mathrm{Dropout}_t$ denotes the dropout mask sampled at trial $t$.
The mean routing vector and the epistemic routing uncertainty are then
defined as
\begin{align}
  \bar{\mathbf{g}}_i &= \frac{1}{T}\sum_{t=1}^T \mathbf{g}_{i,t},
    \label{eq:gate_mean}\\
  u_i &= \sum_{m=1}^3 \mathrm{Var}_{t=1,\ldots,T}
    \bigl[g_{i,m,t}\bigr],
    \label{eq:uncertainty}
\end{align}
where $g_{i,m,t}$ is the $m$-th component of $\mathbf{g}_{i,t}$.

The mean gate values $\bar{\mathbf{g}}_i \in [0,1]^3$ are thresholded
at $\tau = 0.3$, and experts below the threshold are pruned. The fused
latent representation is computed as
\begin{equation}
  \mathbf{z}_i = \sum_{m=1}^3 \tilde{g}_{i,m}\,
    f_m(\mathbf{e}'_{m,i}),
  \quad
  \tilde{\mathbf{g}}_i =
  \frac{\bar{\mathbf{g}}_i \odot
    \mathbf{1}[\bar{\mathbf{g}}_i \ge \tau]}
    {\left\|\bar{\mathbf{g}}_i \odot
    \mathbf{1}[\bar{\mathbf{g}}_i \ge \tau]\right\|_1},
\end{equation}
where $f_m$ denotes the modality-specific expert network for modality $m$.

The scalar $u_i$ in Eq.~\ref{eq:uncertainty} is defined as the
\textbf{epistemic routing uncertainty} for spot $i$, because it aggregates
the variance of the modality-selection weights across MC-Dropout samples. The vector $\bar{\mathbf{g}}_i$ is defined as the
\textbf{modality-routing} vector and is used as an interpretable proxy
for the relative contribution of RNA, ADT, and histology to the fused
representation. Both $u_i$ and $\bar{\mathbf{g}}_i$ are provided as interpretable model
outputs and used by Task~C.
Collecting the mean routing vectors across spots yields the modality-routing
matrix $\mathbf{G}$ introduced in Task~A, whose $i$-th row is
$\bar{\mathbf{g}}_i$.

\subsection{Component 5: ClusteringHead (Task~A)}

The L2-normalized latent representation
$\hat{\mathbf{z}}_i = \mathbf{z}_i /\|\mathbf{z}_i\|_2$ is softly
assigned to $K$ learnable centroids
$\{\boldsymbol{\mu}_k \in \R^d\}_{k=1}^K$ using a Student's
$t$-kernel with one degree of freedom:
\begin{equation}
  q_{ik} =
  \frac{(1 + \|\hat{\mathbf{z}}_i - \boldsymbol{\mu}_k\|_2^2)^{-1}}
       {\sum_{k'=1}^K
       (1 + \|\hat{\mathbf{z}}_i - \boldsymbol{\mu}_{k'}\|_2^2)^{-1}}.
\end{equation}
The predicted domain for spot $i$ is
$\hat{k}_i = \arg\max_k q_{ik}$, and the
\textbf{assignment confidence} is defined as
$c_i = \max_k q_{ik}$.

The clustering objective is a self-sharpening KL divergence,
$\mathcal{L}_\mathrm{cluster} = \mathrm{KL}(\mathbf{P}\|\mathbf{Q})$,
where the auxiliary target distribution is defined as
\begin{equation}
  p_{ik} =
  \frac{q_{ik}^2/f_k}{\sum_{k'=1}^K q_{ik'}^2/f_{k'}},
  \qquad
  f_k = \sum_i q_{ik}.
\end{equation}
The centroids are warm-started by fitting $k$-means on the latent
representations before the clustering loss is activated at warm-up
epoch 50.

\subsection{Component 6: CellTypeHead and ImputationHeads}

A two-layer MLP CellTypeHead produces per-spot class logits for
the semi-supervised auxiliary objective (Task~B,
Section~\ref{sec:taskb}).
Three ImputationHeads, one for each modality, reconstruct randomly masked
entries within modality-specific feature vectors from the fused latent
representation $\mathbf{Z}$. This imputation task provides a
self-supervised reconstruction objective that regularizes $\mathbf{Z}$ by
forcing it to preserve information useful for recovering partially
observed modalities. Consequently, the learned representation becomes
more robust to incomplete, noisy, or sparsely measured features.

\subsection{Training objective}

OmicSync is trained end-to-end with the following objective:
\begin{equation}
\begin{aligned}
  \mathcal{L}_{\text{OmicSync}}
  &= \lambda_r\mathcal{L}_\mathrm{recon}
   + \lambda_c(t)\mathcal{L}_\mathrm{contrast}
   + \lambda_k\mathcal{L}_\mathrm{cluster} \\
  &\quad
   + \lambda_b\mathcal{L}_\mathrm{celltype}
   + \lambda_p\mathcal{L}_\mathrm{impute}
   + \lambda_u\mathcal{L}_\mathrm{unc}.
\end{aligned}
\end{equation}
where $\mathcal{L}_\mathrm{recon}$ denotes per-modality MSE
reconstruction; $\mathcal{L}_\mathrm{contrast}$ is a topology-aware
InfoNCE loss across the three cross-modal embeddings using precomputed
spatial $k$-hop exclusion masks; $\mathcal{L}_\mathrm{cluster}$ is the
self-sharpening clustering loss in Task~A; $\mathcal{L}_\mathrm{celltype}$
is the cross-entropy loss evaluated on labelled spots only;
$\mathcal{L}_\mathrm{impute}$ is the MSE reconstruction loss for the
masked modality; and $\mathcal{L}_\mathrm{unc}$ is a light regulariser
on the mean routing uncertainty.

\paragraph{Adaptive spatial exclusion radius.}
For the topology-aware InfoNCE loss, we define $n_\mathrm{hops}$ as the
spatial exclusion radius: spots within $n_\mathrm{hops}$ steps on the
spatial graph are excluded from the negative set, so that nearby spots
are not incorrectly treated as negatives. Before training, OmicSync
measures the average cosine similarity between each spot and its direct
spatial neighbors in RNA space. Homogeneous tissue, defined by an average
neighbor similarity greater than $0.6$ (e.g., tonsil germinal centers),
uses $n_\mathrm{hops}=2$ to apply a broader exclusion neighborhood.
Heterogeneous tissue, such as breast-cancer tumor--stroma boundaries,
uses $n_\mathrm{hops}=1$ to avoid excluding too many transcriptionally
distinct nearby spots from contrastive learning.

\paragraph{Contrastive weight decay.}
The contrastive weight $\lambda_c(t)$ is held at its base value for the
first half of training and then decays linearly to a floor of
$\lambda_c/3$. This schedule allows the contrastive loss to promote
local spatial consistency early in training, while allowing the clustering
and reconstruction objectives to play a larger role later. As a result,
the contrastive term helps establish spatially coherent representations
without overly constraining the RNA-defined domain boundaries used for
clustering evaluation.

\section{Task B: Cell-type regularisation and spatial prediction maps}
\label{sec:taskb}

The spatial multi-omics datasets used in this study do not provide
external ground-truth cell-type annotations matched across the RNA,
protein, and histology modalities. We therefore treat the CellTypeHead
not as an independent supervised cell-type annotation benchmark, but as
a \textbf{semi-supervised regularisation} branch.

Leiden community detection on the RNA representation is used to obtain
coarse pseudo-labels. We use 70\% of these pseudo-labels as auxiliary
training targets and hold out the remaining 30\% to avoid using all
pseudo-labels directly for optimisation. The auxiliary cross-entropy loss $\mathcal{L}_\mathrm{celltype}$ uses
Leiden-derived pseudo-labels as partial training targets and provides a
structural prior that encourages the shared latent space to preserve
local pseudo-cell-type organisation.

After training, the CellTypeHead is applied to all spots to generate
spatial pseudo-cell-type prediction maps. Accordingly, we report Task~B
qualitatively through these prediction maps rather than claiming it as an
independent supervised cell-type annotation benchmark.

\section{Task C: Coupled Reasoning}
\label{sec:taskc}
Task~C aims to translate OmicSync's quantitative spot-level outputs into
human-readable explanations while preserving faithfulness to the underlying
model evidence.  Rather than introducing an additional trainable component,
Task~C operates as a post-hoc evidence-coupled reasoning module: it receives
structured outputs from Task~A, including domain assignment, assignment
confidence, modality-routing weights, routing uncertainty, marker evidence,
and local neighbourhood composition, and uses them as the sole evidence source
for explanation generation. This coupling is designed to make the explanations
auditable, so that each generated justification can be traced back to the
model-derived evidence supporting the corresponding spatial-domain assignment.

\subsection{The coupling mechanism}

Task~C is a post-hoc evidence-constrained reasoning module. 
It does not backpropagate into the clustering objective; instead, it
uses only model-derived signals produced by Task~A.
The module is evidence-constrained in the sense that the language model is
provided only with the extracted evidence dictionary and is instructed not to
introduce unsupported biological entities or claims.
For each explained spot $i$, an evidence-extraction routine assembles:
\begin{enumerate}
  \item The predicted domain $\hat{k}_i = \arg\max_k q_{ik}$ and
        assignment confidence $c_i = \max_k q_{ik}$;
  \item The runner-up domain
        $\hat{k}_i^{(2)} = \arg\max_{k \neq \hat{k}_i} q_{ik}$ and
        the confidence margin
        $\Delta_i = q_{i\hat{k}_i} - q_{i\hat{k}_i^{(2)}}$;
  \item The modality-routing vector $\bar{\mathbf{g}}_i$ and the
        dominant modality $\arg\max_m \bar{g}_{i,m}$;
  \item The epistemic routing uncertainty $u_i$;
  \item The top evidence features associated with the dominant modality;
  \item The composition of the five-nearest-spot neighbourhood.
\end{enumerate}
This evidence dictionary is the sole structured input to the language
model. The reasoning module is therefore evidence-constrained: the
language model is instructed not to introduce gene names, protein names,
spot identifiers, citations, or biological claims that are absent from
the supplied evidence.

\subsection{Five reasoning strategies}

The same evidence dictionary is rendered into five prompt templates,
each addressing a distinct explanatory question:

\noindent\textbf{Standard.}
Direct evidence-to-conclusion reasoning: given the dominant modality,
top markers, neighbourhood composition, confidence margin, and uncertainty
tier, why is spot $i$ assigned to domain $\hat{k}_i$?

\noindent\textbf{Stepwise.}
A structured five-step rationale: (1) molecular or image-derived evidence,
(2) modality-weighting rationale, (3) spatial-context support or
contradiction, (4) confidence and uncertainty interpretation, and
(5) final conclusion.

\noindent\textbf{Counterfactual.}
Three targeted counterfactual questions: if the top evidence features
were absent, if the dominant modality were unavailable, or if the local
neighbourhood shifted toward the runner-up domain, would the prediction
still be supported?

\noindent\textbf{Contrastive.}
Why was domain $\hat{k}_i$ selected over the runner-up domain
$\hat{k}_i^{(2)}$, given the top evidence features, dominant modality,
and confidence margin?

\noindent\textbf{Uncertainty.}
Given the epistemic uncertainty tier (LOW/MEDIUM/HIGH), the modality
routing distribution, and the spatial context, should the assignment be
trusted for downstream analysis?

\subsection{Language backbone and faithfulness constraints}

Justifications are generated by Llama-3.3-70B accessed through the
Groq inference API.
A strict system prompt instructs the model to use only the evidence
provided in the prompt and forbids introducing gene names, protein names,
spot indices, citations, or biological facts not present in the supplied
evidence.
This design constrains
the generated explanations to the model-derived evidence used for the
corresponding spot-level audit.

\subsection{Reasoning quality metrics}

Three automatically computed faithfulness metrics evaluate each
justification against the supplied evidence:

\noindent\textbf{Grounding rate (GR).}
For text-based marker evidence, GR measures the fraction of supplied
named marker genes or proteins that are mentioned in the justification:
\begin{equation}
  \mathrm{GR}_i =
  \frac{
  \left|
  \mathcal{M}_i \cap \mathcal{J}_i
  \right|
  }{
  \left|
  \mathcal{M}_i
  \right|
  },
\end{equation}
where $\mathcal{M}_i$ is the set of named marker features supplied in
the evidence dictionary and $\mathcal{J}_i$ is the set of marker names
detected in the generated justification. Spots whose dominant evidence
comes from unnamed image features are excluded from GR computation.

\noindent\textbf{Confidence alignment (CA).}
CA evaluates whether the confidence and hedging language in the
justification is consistent with the epistemic uncertainty tier. HIGH
uncertainty should lead to cautious language, whereas LOW uncertainty
should lead to more confident language.

\noindent\textbf{Spatial consistency (SC).}
SC evaluates whether the generated justification correctly references
the dominant neighbourhood composition supplied in the evidence
dictionary.

\section{OmicSync-R: Reasoning-Guided Training via Policy Gradient}
\label{sec:omicsync_r}

In the base OmicSync framework, the reasoning module operates post hoc:
it consumes model-derived evidence but provides no feedback to the
clustering objective. We further consider \textbf{OmicSync-R}, a
reasoning-guided variant in which automatically computed reasoning
quality scores are used as reward signals for the ClusteringHead through
the REINFORCE policy-gradient estimator~\citep{williams1992simple}.
The language model and text-scoring procedure are treated as black-box
components. Therefore, no gradient is taken through the generated
justification; instead, the scalar reward is used to weight the
log-probability of the sampled cluster assignment.

\subsection{Formulation}

Let $\mathbf{q}_i \in [0,1]^K$ denote the soft assignment distribution
produced by the ClusteringHead for spot $i$. At each reasoning-update
step, a candidate domain label $a_i$ is sampled from this distribution:
\begin{equation}
  a_i \sim \mathrm{Categorical}(\mathbf{q}_i).
\end{equation}
An evidence dictionary conditioned on the sampled domain $a_i$ is then
assembled and passed to Llama-3.3-70B to generate a justification $J_i$.
The justification is scored using the automatically computed reasoning
quality metrics:
\begin{equation}
  R_i = w_1 \mathrm{GR}_i
      + w_2 \mathrm{CA}_i
      + w_3 \mathrm{SC}_i
      \in [0,1],
  \label{eq:reward}
\end{equation}
where $w_1 = 0.5$, $w_2 = 0.3$, and $w_3 = 0.2$.
Grounding rate receives the highest weight because it is the most
directly verifiable faithfulness criterion: it measures whether the
generated justification refers to the marker evidence supplied to the
language model.

A domain-stratified sample $\mathcal{S}_R$, with
$|\mathcal{S}_R|=10$, is drawn every $N_R=25$ epochs after the warm-up
period. The mean reward at reasoning-update epoch $t$ is
\begin{equation}
  \bar{R}_t =
  \frac{1}{|\mathcal{S}_R|}
  \sum_{i \in \mathcal{S}_R} R_i.
\end{equation}
An exponential moving average baseline is used to reduce gradient
variance:
\begin{equation}
  b_t = \alpha b_{t-1} + (1-\alpha)\bar{R}_t,
  \qquad
  \alpha = 0.9,\quad b_0 = 0.5.
\end{equation}
The REINFORCE loss applied to the assignment distribution is
\begin{equation}
  \mathcal{L}_\mathrm{reinforce} =
  -\frac{1}{|\mathcal{S}_R|}
  \sum_{i \in \mathcal{S}_R}
  \mathrm{sg}\!\left(R_i - b_t\right)
  \log q_{i,a_i},
  \label{eq:reinforce}
\end{equation}
where $\mathrm{sg}(\cdot)$ denotes stop-gradient, indicating that the
reward and baseline are treated as constants during backpropagation.
A positive advantage $(R_i-b_t)>0$ increases the probability of the
sampled assignment that produced a well-grounded justification, whereas
a negative advantage decreases the probability of assignments whose
justifications are poorly grounded, poorly calibrated, or spatially
inconsistent.

\subsection{Extended training objective}

The OmicSync-R objective augments the base OmicSync training loss:
\begin{equation}
  \mathcal{L}_{\text{OmicSync-R}}
  =
  \mathcal{L}_{\text{OmicSync}}
  + \lambda_R \mathcal{L}_\mathrm{reinforce}.
\end{equation}
Here, $\mathcal{L}_{\text{OmicSync}}$ denotes the base OmicSync objective,
and $\lambda_R$ controls the contribution of the reasoning-guided
policy-gradient term. We set $\lambda_R=0.02$, substantially smaller than
the clustering weight $\lambda_k=1.0$ (contained in $\mathcal{L}_{\text{OmicSync}}$), so that geometric domain separation
remains the primary objective and reasoning coherence acts as a secondary
regulariser.
The REINFORCE term is non-zero only at reasoning-update
epochs; all other training steps use the base OmicSync objective.
Base OmicSync is recovered as the special case $\lambda_R=0$.

\subsection{Training procedure}

Algorithm~\ref{alg:omicsync_r} summarises the OmicSync-R training loop.
The REINFORCE update is activated only after the warm-up period
($t_\mathrm{warm}=50$), consistent with the clustering loss schedule, to
avoid using unstable early-training explanations as reward signals.
With $|\mathcal{S}_R|=10$, $N_R=25$, and a 600-epoch training schedule,
the LLM is queried 10 times per reasoning-update epoch.
Updates occur at epochs $50,75,\ldots,575$, yielding 22 reasoning-update
steps and 220 LLM queries in total.
Because OmicSync-R requires repeated LLM-based explanation generation and
reward evaluation during training, this reasoning-guided variant is more
resource-intensive than base OmicSync. We therefore evaluate OmicSync-R on
the Human Breast Cancer dataset as a representative case study rather than
across all benchmarks.

\begin{algorithm}[t]
\caption{OmicSync-R: reasoning-guided training loop}
\label{alg:omicsync_r}
\begin{algorithmic}[1]
\Require Modalities $\{\mathbf{X}_1,\mathbf{X}_2,\mathbf{X}_3,
         \mathbf{C}\}$, LLM generator, scoring routine,
         $N_R, \lambda_R, \alpha$
\State Initialise OmicSync model
\State $b_0 \leftarrow 0.5$
\For{$t = 1, \ldots, T$}
  \State Forward pass: obtain $\mathbf{Z}, \mathbf{Q},
         \{\bar{\mathbf{g}}_i\}, \{u_i\}$
  \If{$t = t_\mathrm{warm}$}
    \State Warm-start centroids using $k$-means on the current latent
           representations
  \EndIf
  \State Compute base OmicSync losses according to the training schedule
  \State $\mathcal{L}_\mathrm{reinforce} \leftarrow 0$
  \If{$t \ge t_\mathrm{warm}$ \textbf{and} $t \bmod N_R = 0$}
    \State Draw a domain-stratified sample $\mathcal{S}_R$,
           $|\mathcal{S}_R|=10$
    \For{each $i \in \mathcal{S}_R$}
      \State $a_i \sim \mathrm{Categorical}(\mathbf{q}_i)$
      \State Assemble evidence dictionary conditioned on domain $a_i$
      \State Query LLM and obtain justification $J_i$
      \State Compute reward $R_i$ using Eq.~\eqref{eq:reward}
    \EndFor
    \State $\bar{R}_t \leftarrow
           \frac{1}{|\mathcal{S}_R|}\sum_{i \in \mathcal{S}_R} R_i$
    \State $b_t \leftarrow \alpha b_{t-1} + (1-\alpha)\bar{R}_t$
    \State Compute $\mathcal{L}_\mathrm{reinforce}$ using
           Eq.~\eqref{eq:reinforce}
  \EndIf
  \State $\mathcal{L} \leftarrow \mathcal{L}_{\text{OmicSync}}
         + \lambda_R\mathcal{L}_\mathrm{reinforce}$
  \State Backward pass and Adam update
\EndFor
\end{algorithmic}
\end{algorithm}

\section{Experiments}
\label{sec:experiments}

\subsection{Datasets}

We evaluate OmicSync on four 10x Genomics CytAssist FFPE Protein
Expression datasets: \textbf{Human Tonsil} (4,194 spots),
\textbf{Human Breast Cancer} (3,786 spots),
\textbf{Human Glioblastoma} (3,460 spots), and
\textbf{Human Tonsil with Add-on Antibodies} (3,512 spots).
Each dataset provides a feature-barcode matrix containing RNA
(Gene Expression) and protein (Antibody Capture) modalities, spot
pixel coordinates, and an H\&E image. Histology embeddings are
extracted with UNI as described in Section~\ref{sec:method}.

\subsection{Baselines}

We compare OmicSync against four representative state-of-the-art
methods for spatial multi-omics domain discovery:
GROVER~\citep{grover2024}, MISO~\citep{miso2024},
SpatialGlue~\citep{spatialglue2024}, and COSMOS~\citep{cosmos2024}.
These baselines were selected because they cover diverse strategies for
spatial multi-omics representation learning, including graph-based
integration, multi-scale modelling, cross-modal attention, and
contrastive multi-view learning. Together, they provide a broad
comparison against established multimodal fusion and clustering
frameworks.

\subsection{Implementation details}

OmicSync is implemented in PyTorch and trained on an NVIDIA H100 GPU.
We use latent dimension $d=64$, $K \in \{6,7,8,9,10\}$, 50 warm-up
epochs, and 600 total training epochs. The learning rate is initialized
at $10^{-4}$ and decayed to $10^{-6}$ using cosine annealing. We use
the Adam optimiser with gradient norm clipping at 1.0. The loss weights
are set to $\lambda_r=1.0$, $\lambda_c=1.5$, $\lambda_k=1.0$,
$\lambda_b=1.0$, $\lambda_p=0.5$, and $\lambda_u=0.01$.
The CrossModalTransformer uses 4 attention heads and 2 layers. The MoE
threshold is set to $\tau=0.3$, MC-Dropout uses $T=10$ stochastic
samples, and the imputation masking probability is 0.15.

Deterministic seeding is applied to Python, NumPy, PyTorch, CUDA, and
cuDNN to improve reproducibility and reduce seed-dependent variation.
For OmicSync-R, we use $\lambda_R=0.02$, $N_R=25$, $\alpha=0.9$, and
$|\mathcal{S}_R|=10$. Because OmicSync-R requires repeated LLM-based
explanation generation and reward evaluation during training, this
reasoning-guided variant is more resource-intensive than base OmicSync;
we therefore evaluate OmicSync-R on the Human Breast Cancer dataset as a
representative case study.

\subsection{Evaluation metrics}

Task~A is evaluated using adjusted Rand index (ARI), normalized mutual
information (NMI), Fowlkes--Mallows index (FMI), silhouette coefficient
(SilC), adjusted mutual information (AMI), pairwise Jaccard index,
Calinski--Harabasz index (CHI), Purity, and Davies--Bouldin index (DBI).
Because the original datasets do not provide expert-annotated
spot-level spatial domains, we adopt the curated pseudo-reference labels
used by GROVER~\citep{grover2024} for external clustering evaluation.
These labels are derived from the available molecular modalities and
provide a consistent reference partition for comparing all methods.
ARI, NMI, FMI, AMI, pairwise Jaccard index, and Purity are computed
against these pseudo-reference labels, whereas SilC, CHI, and DBI
evaluate the intrinsic geometry of the predicted clusters.

For fair comparison with embedding-based baselines, $k$-means is applied
to the learned latent representation $\mathbf{Z}$ for
$K \in \{6,7,8,9,10\}$. The same range of $K$ and the same evaluation
protocol are used for all methods, and the best result over this shared
range is reported for each method.
Separately, the trained ClusteringHead produces soft assignment
probabilities $\mathbf{Q}$, which are used for confidence estimation,
runner-up analysis, and Task~C reasoning. Task~C is evaluated using
grounding rate (GR), confidence alignment (CA), and spatial consistency
(SC), as defined in Section~\ref{sec:taskc}.

\section{Results}
\label{sec:results}

\subsection{Task A: Clustering performance}

Table~\ref{tab:clustering_results} reports clustering performance on
all four datasets. OmicSync achieves the best average rank on Human
Tonsil (1.44), Human Glioblastoma (1.78), and Human Tonsil with Add-on
Antibodies (1.22), and the second-best average rank on Human Breast
Cancer (2.33). It obtains the best ARI on all four datasets
(46.81, 45.73, 45.74, and 53.80, respectively), with low variation
across repeated runs. 
OmicSync also achieves the highest CHI values by a large margin, for
example 4,899 versus 2,494 on Human Tonsil, 2,990 versus 2,436 on Human
Breast Cancer, 7,904 versus 1,430 on Human Glioblastoma, and 5,990
versus 3,979 on Human Tonsil with Add-on Antibodies. 
These results indicate that OmicSync learns latent representations whose cluster assignments better align with the curated reference partitions, as reflected by ARI, while also exhibiting improved intrinsic cluster separation and compactness on most datasets, as reflected by CHI, SilC, and DBI.

\paragraph{Where OmicSync wins and why.}
The largest ARI improvement is observed on Human Tonsil with Add-on
Antibodies, where OmicSync achieves an ARI of 53.80, compared with 46.5
for the strongest baseline, GROVER. This improvement suggests that the
richer antibody panel may provide a stronger protein signal that can be
effectively exploited by the CrossModalTransformer and the
uncertainty-aware MoE fusion mechanism. The adaptive
$n_\mathrm{hops}$ mechanism also operates differently across datasets:
the tonsil datasets use $n_\mathrm{hops}=2$ because of their more
homogeneous germinal-centre structure and higher average spatial RNA
similarity, whereas Human Breast Cancer uses $n_\mathrm{hops}=1$ to
account for heterogeneous tumour--stroma boundaries. This
dataset-adaptive exclusion strategy is designed to avoid treating nearby
spots in homogeneous regions as negative pairs, while retaining greater
sensitivity to transcriptionally distinct neighbouring spots in
heterogeneous regions.

\paragraph{Purity trade-off on Tonsil.}
On Human Tonsil, OmicSync achieves lower Purity (57.71) than GROVER
(69.4), despite obtaining better ARI, SilC, CHI, and DBI. This pattern
suggests a trade-off between majority-label dominance within individual
clusters and broader clustering quality. Specifically, OmicSync produces
more compact and better separated clusters, as reflected by its higher
CHI and lower DBI, but does not maximise the proportion of the dominant
reference label within each cluster on this dataset. We therefore
interpret OmicSync's overall advantage using complementary metrics that
capture both agreement with the curated reference labels and intrinsic
latent-space cluster structure, including ARI, CHI, SilC, and DBI.

\begin{table*}[t]
\centering
\renewcommand{\arraystretch}{1.2}
\setlength{\tabcolsep}{3.5pt}
\caption{Clustering performance across spatial proteomics datasets.
\textbf{Bold} = best; \underline{underline} = second best.}
\label{tab:clustering_results}
\begin{adjustbox}{max width=\textwidth}
\tiny
\begin{tabular}{l *{10}{c}}
\toprule
Method
  & ARI$\uparrow$
  & NMI$\uparrow$
  & FMI$\uparrow$
  & SilC$\uparrow$
  & AMI$\uparrow$
  & Jaccard$\uparrow$
  & CHI$\uparrow$
  & Purity$\uparrow$
  & DBI$\downarrow$
  & Rank$\downarrow$ \\
\midrule
\multicolumn{11}{c}{\textit{Human Tonsil}} \\
\midrule
GROVER
  & $45.2{\scriptstyle\pm7.8}$
  & $54.3{\scriptstyle\pm9.9}$
  & $\mathbf{54.1{\scriptstyle\pm6.8}}$
  & $\underline{31.6{\scriptstyle\pm3.9}}$
  & $\underline{54.2{\scriptstyle\pm10.1}}$
  & $\underline{37.3{\scriptstyle\pm6.6}}$
  & $2494{\scriptstyle\pm286}$
  & $\mathbf{69.4{\scriptstyle\pm5.4}}$
  & $\underline{139.8{\scriptstyle\pm10.5}}$
  & $\underline{1.78}$ \\
MISO
  & $41.3{\scriptstyle\pm6.7}$
  & $51.2{\scriptstyle\pm4.6}$
  & $52.5{\scriptstyle\pm4.3}$
  & $7.0{\scriptstyle\pm1.6}$
  & $51.2{\scriptstyle\pm4.6}$
  & $35.4{\scriptstyle\pm3.8}$
  & $244{\scriptstyle\pm15}$
  & $64.2{\scriptstyle\pm5.5}$
  & $203.4{\scriptstyle\pm14.8}$
  & $4.00$ \\
SpatialGlue
  & $\underline{43.3{\scriptstyle\pm6.7}}$
  & $\underline{53.9{\scriptstyle\pm8.9}}$
  & $52.4{\scriptstyle\pm6.1}$
  & $23.8{\scriptstyle\pm3.2}$
  & $53.9{\scriptstyle\pm8.9}$
  & $35.3{\scriptstyle\pm5.6}$
  & $1064{\scriptstyle\pm124}$
  & $\underline{68.7{\scriptstyle\pm5.0}}$
  & $159.6{\scriptstyle\pm7.0}$
  & $3.22$ \\
COSMOS
  & $19.8{\scriptstyle\pm6.7}$
  & $27.9{\scriptstyle\pm6.0}$
  & $32.3{\scriptstyle\pm6.6}$
  & $20.0{\scriptstyle\pm0.7}$
  & $27.6{\scriptstyle\pm5.6}$
  & $19.3{\scriptstyle\pm4.9}$
  & $937{\scriptstyle\pm100}$
  & $49.9{\scriptstyle\pm9.0}$
  & $157.8{\scriptstyle\pm4.2}$
  & $4.56$ \\
OmicSync
  & $\mathbf{46.81{\scriptstyle\pm0.02}}$
  & $\mathbf{55.52{\scriptstyle\pm0.03}}$
  & $\underline{52.98{\scriptstyle\pm0.02}}$
  & $\mathbf{32.77{\scriptstyle\pm0.01}}$
  & $\mathbf{55.90{\scriptstyle\pm0.03}}$
  & $\mathbf{38.74{\scriptstyle\pm0.01}}$
  & $\mathbf{4899{\scriptstyle\pm0.3}}$
  & $57.71{\scriptstyle\pm0.02}$
  & $\mathbf{124.74{\scriptstyle\pm0.16}}$
  & $\mathbf{1.44}$ \\
\midrule
\multicolumn{11}{c}{\textit{Human Breast Cancer}} \\
\midrule
GROVER
  & $\underline{44.1{\scriptstyle\pm10.7}}$
  & $52.4{\scriptstyle\pm8.7}$
  & $\mathbf{53.9{\scriptstyle\pm8.6}}$
  & $\mathbf{36.3{\scriptstyle\pm7.7}}$
  & $\underline{52.3{\scriptstyle\pm8.6}}$
  & $\mathbf{37.3{\scriptstyle\pm8.1}}$
  & $\underline{2436{\scriptstyle\pm385}}$
  & $\underline{64.8{\scriptstyle\pm9.9}}$
  & $\mathbf{139.6{\scriptstyle\pm13.8}}$
  & $\mathbf{1.67}$ \\
MISO
  & $37.5{\scriptstyle\pm3.0}$
  & $47.9{\scriptstyle\pm2.0}$
  & $49.8{\scriptstyle\pm3.0}$
  & $11.0{\scriptstyle\pm0.6}$
  & $47.7{\scriptstyle\pm2.0}$
  & $32.7{\scriptstyle\pm2.7}$
  & $289{\scriptstyle\pm21}$
  & $56.7{\scriptstyle\pm3.6}$
  & $211.5{\scriptstyle\pm10.7}$
  & $4.22$ \\
SpatialGlue
  & $43.0{\scriptstyle\pm6.9}$
  & $\underline{53.0{\scriptstyle\pm5.1}}$
  & $\underline{52.1{\scriptstyle\pm6.1}}$
  & $20.2{\scriptstyle\pm0.8}$
  & $\mathbf{53.5{\scriptstyle\pm4.8}}$
  & $35.2{\scriptstyle\pm6.0}$
  & $1175{\scriptstyle\pm135}$
  & $\mathbf{67.2{\scriptstyle\pm5.0}}$
  & $172.2{\scriptstyle\pm3.3}$
  & $2.56$ \\
COSMOS
  & $25.6{\scriptstyle\pm2.2}$
  & $36.5{\scriptstyle\pm3.5}$
  & $37.0{\scriptstyle\pm1.8}$
  & $24.8{\scriptstyle\pm0.8}$
  & $36.3{\scriptstyle\pm3.5}$
  & $22.7{\scriptstyle\pm1.6}$
  & $1226{\scriptstyle\pm106}$
  & $54.5{\scriptstyle\pm2.9}$
  & $\underline{143.4{\scriptstyle\pm2.6}}$
  & $4.22$ \\
OmicSync
  & $\mathbf{45.73{\scriptstyle\pm2.21}}$
  & $\mathbf{54.1{\scriptstyle\pm0.95}}$
  & $51.75{\scriptstyle\pm2.35}$
  & $\underline{28.92{\scriptstyle\pm1.25}}$
  & $43.72{\scriptstyle\pm0.95}$
  & $\underline{35.70{\scriptstyle\pm1.60}}$
  & $\mathbf{2990{\scriptstyle\pm8}}$
  & $56.78{\scriptstyle\pm1.27}$
  & $178.76{\scriptstyle\pm6.96}$
  & $\underline{2.33}$ \\
\midrule
\multicolumn{11}{c}{\textit{Human Glioblastoma}} \\
\midrule
GROVER
  & $\underline{40.8{\scriptstyle\pm6.6}}$
  & $\underline{53.9{\scriptstyle\pm4.1}}$
  & $51.6{\scriptstyle\pm4.6}$
  & $22.6{\scriptstyle\pm1.1}$
  & $\underline{53.8{\scriptstyle\pm3.8}}$
  & $\underline{34.1{\scriptstyle\pm4.8}}$
  & $1413{\scriptstyle\pm111}$
  & $\underline{71.9{\scriptstyle\pm3.1}}$
  & $157.0{\scriptstyle\pm3.8}$
  & $\underline{2.67}$ \\
MISO
  & $43.5{\scriptstyle\pm6.9}$
  & $49.2{\scriptstyle\pm2.2}$
  & $\mathbf{55.5{\scriptstyle\pm7.0}}$
  & $9.6{\scriptstyle\pm2.9}$
  & $49.0{\scriptstyle\pm2.2}$
  & $\mathbf{38.4{\scriptstyle\pm7.2}}$
  & $421{\scriptstyle\pm47}$
  & $65.3{\scriptstyle\pm7.5}$
  & $235.8{\scriptstyle\pm10.3}$
  & $3.44$ \\
SpatialGlue
  & $40.1{\scriptstyle\pm7.6}$
  & $53.8{\scriptstyle\pm7.3}$
  & $50.9{\scriptstyle\pm5.5}$
  & $23.4{\scriptstyle\pm0.5}$
  & $\underline{53.8{\scriptstyle\pm7.3}}$
  & $33.4{\scriptstyle\pm5.4}$
  & $\underline{1430{\scriptstyle\pm133}}$
  & $\mathbf{72.3{\scriptstyle\pm3.6}}$
  & $157.2{\scriptstyle\pm3.9}$
  & $2.89$ \\
COSMOS
  & $32.0{\scriptstyle\pm6.9}$
  & $48.6{\scriptstyle\pm4.3}$
  & $44.2{\scriptstyle\pm4.5}$
  & $\underline{25.8{\scriptstyle\pm2.4}}$
  & $48.4{\scriptstyle\pm4.2}$
  & $28.0{\scriptstyle\pm4.1}$
  & $1325{\scriptstyle\pm92}$
  & $67.8{\scriptstyle\pm3.9}$
  & $\underline{137.4{\scriptstyle\pm9.0}}$
  & $3.89$ \\
OmicSync
  & $\mathbf{45.74{\scriptstyle\pm0.74}}$
  & $\mathbf{56.19{\scriptstyle\pm0.69}}$
  & $\underline{52.23{\scriptstyle\pm0.86}}$
  & $\mathbf{26.72{\scriptstyle\pm0.84}}$
  & $\mathbf{56.03{\scriptstyle\pm0.69}}$
  & $29.31{\scriptstyle\pm0.59}$
  & $\mathbf{7904{\scriptstyle\pm64}}$
  & $67.75{\scriptstyle\pm1.10}$
  & $\mathbf{134.4{\scriptstyle\pm4.82}}$
  & $\mathbf{1.78}$ \\
\midrule
\multicolumn{11}{c}{\textit{Human Tonsil with Add-on Antibodies}} \\
\midrule
GROVER
  & $\underline{46.5{\scriptstyle\pm5.6}}$
  & $\underline{59.0{\scriptstyle\pm4.8}}$
  & $55.3{\scriptstyle\pm6.0}$
  & $\mathbf{38.2{\scriptstyle\pm1.2}}$
  & $\underline{58.8{\scriptstyle\pm4.7}}$
  & $38.0{\scriptstyle\pm5.7}$
  & $\underline{3979{\scriptstyle\pm185}}$
  & $\mathbf{70.5{\scriptstyle\pm6.1}}$
  & $\underline{105.8{\scriptstyle\pm2.9}}$
  & $\underline{2.00}$ \\
MISO
  & $44.6{\scriptstyle\pm11.9}$
  & $56.1{\scriptstyle\pm7.6}$
  & $\underline{55.9{\scriptstyle\pm10.4}}$
  & $8.3{\scriptstyle\pm0.5}$
  & $55.9{\scriptstyle\pm7.6}$
  & $\mathbf{38.9{\scriptstyle\pm10.2}}$
  & $357{\scriptstyle\pm33}$
  & $65.5{\scriptstyle\pm11.0}$
  & $217.2{\scriptstyle\pm15.2}$
  & $3.78$ \\
SpatialGlue
  & $45.3{\scriptstyle\pm7.3}$
  & $58.1{\scriptstyle\pm5.7}$
  & $54.1{\scriptstyle\pm7.3}$
  & $21.4{\scriptstyle\pm1.1}$
  & $58.0{\scriptstyle\pm5.8}$
  & $36.9{\scriptstyle\pm6.6}$
  & $1331{\scriptstyle\pm134}$
  & $\mathbf{70.5{\scriptstyle\pm5.9}}$
  & $160.6{\scriptstyle\pm2.9}$
  & $3.00$ \\
COSMOS
  & $24.6{\scriptstyle\pm4.3}$
  & $35.1{\scriptstyle\pm1.0}$
  & $36.4{\scriptstyle\pm5.7}$
  & $18.4{\scriptstyle\pm2.5}$
  & $35.0{\scriptstyle\pm0.9}$
  & $22.1{\scriptstyle\pm4.2}$
  & $1194{\scriptstyle\pm139}$
  & $51.5{\scriptstyle\pm6.8}$
  & $168.8{\scriptstyle\pm5.7}$
  & $4.56$ \\
OmicSync
  & $\mathbf{53.80{\scriptstyle\pm0.05}}$
  & $\mathbf{61.12{\scriptstyle\pm0.06}}$
  & $\mathbf{59.48{\scriptstyle\pm0.04}}$
  & $\underline{34.52{\scriptstyle\pm0.02}}$
  & $\mathbf{59.10{\scriptstyle\pm0.06}}$
  & $\mathbf{39.44{\scriptstyle\pm0.04}}$
  & $\mathbf{5990{\scriptstyle\pm0.1}}$
  & $\underline{66.92{\scriptstyle\pm0.03}}$
  & $\mathbf{102.12{\scriptstyle\pm0.07}}$
  & $\mathbf{1.22}$ \\
\bottomrule
\end{tabular}
\end{adjustbox}
\end{table*}

\subsection{Task B: Pseudo-cell-type regularisation and spatial prediction maps} \label{sec:taskb_results} 
\begin{figure*}[!tpb]
\centering
\begin{subfigure}{0.48\textwidth}
    \centering
    \includegraphics[width=\linewidth,trim=0 0 0 18.3,clip]{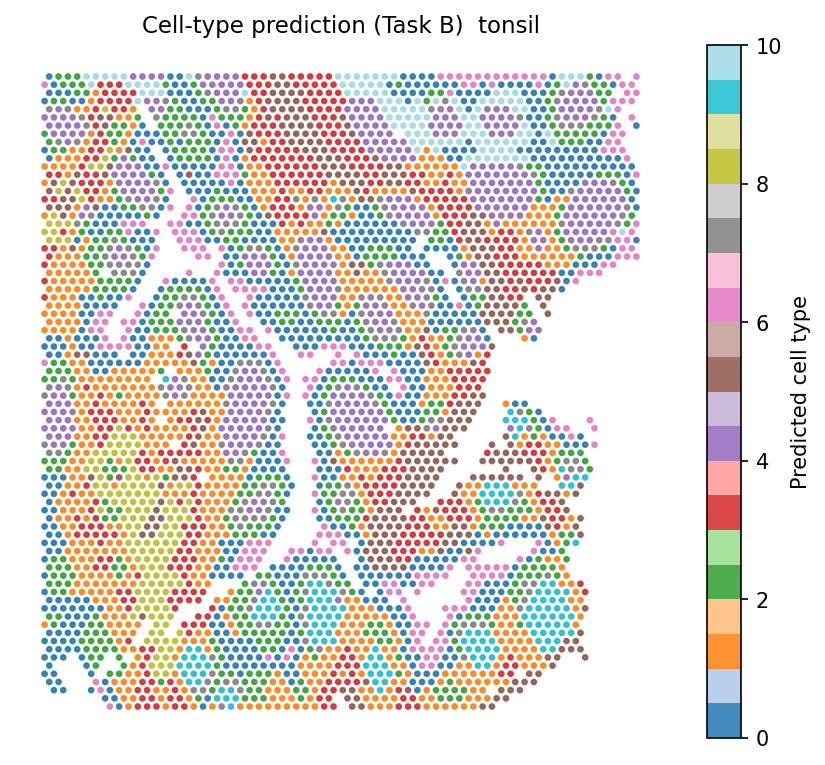}
    \caption{Human Tonsil}
\end{subfigure}
\hfill
\begin{subfigure}{0.48\textwidth}
    \centering
    \includegraphics[width=\linewidth,trim=0 0 0 18.3,clip]{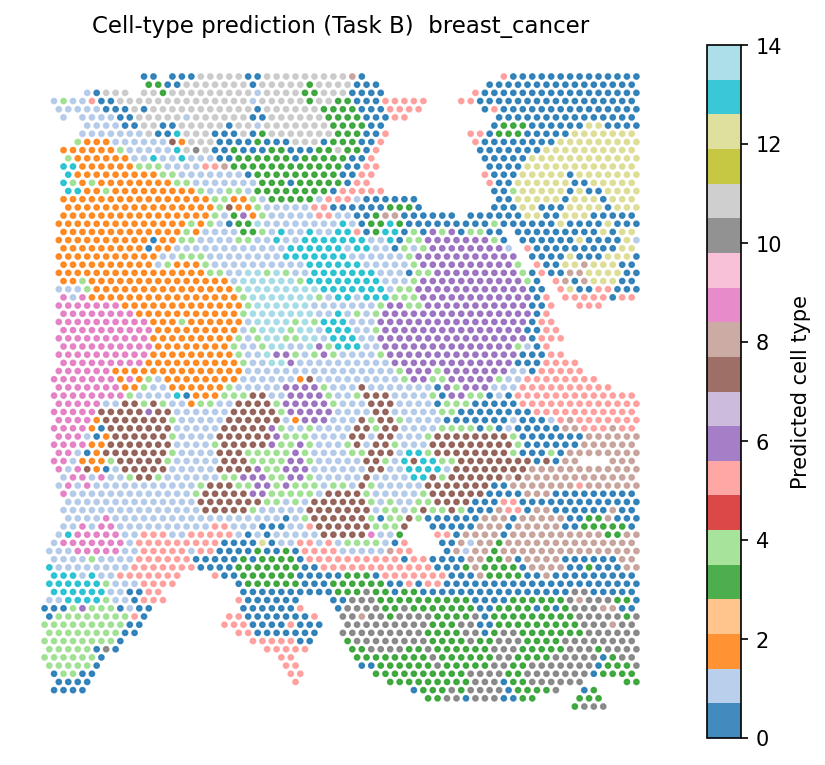}
    \caption{Human Breast Cancer}
\end{subfigure}
\vspace{0.5em}
\begin{subfigure}{0.48\textwidth}
    \centering
    \includegraphics[width=\linewidth,trim=0 0 0 18,clip]{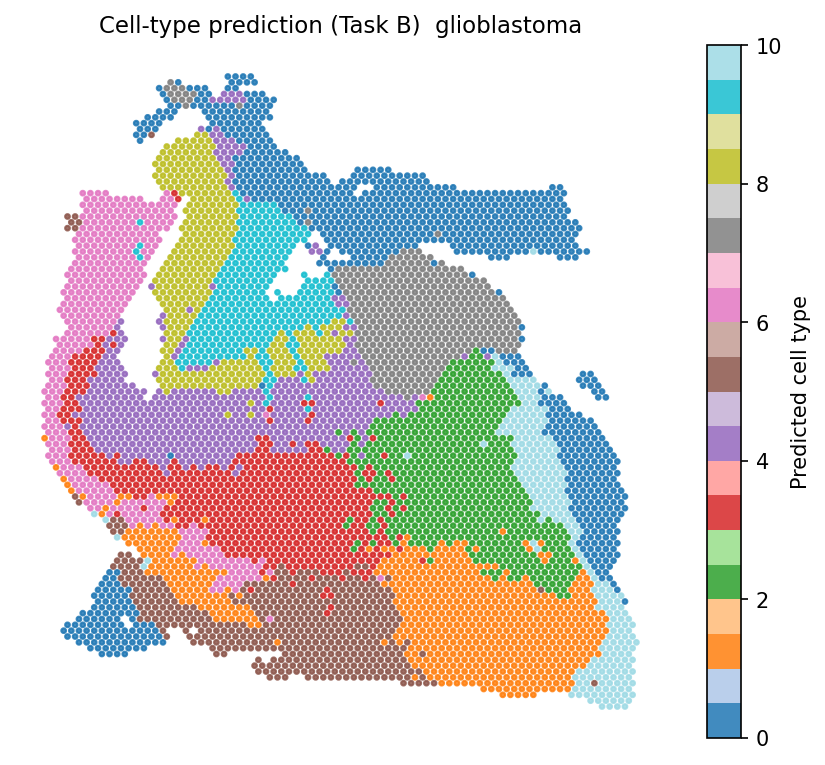}
    \caption{Human Glioblastoma}
\end{subfigure}
\hfill
\begin{subfigure}{0.48\textwidth}
    \centering
    \includegraphics[width=\linewidth,trim=0 0 0 18,clip]{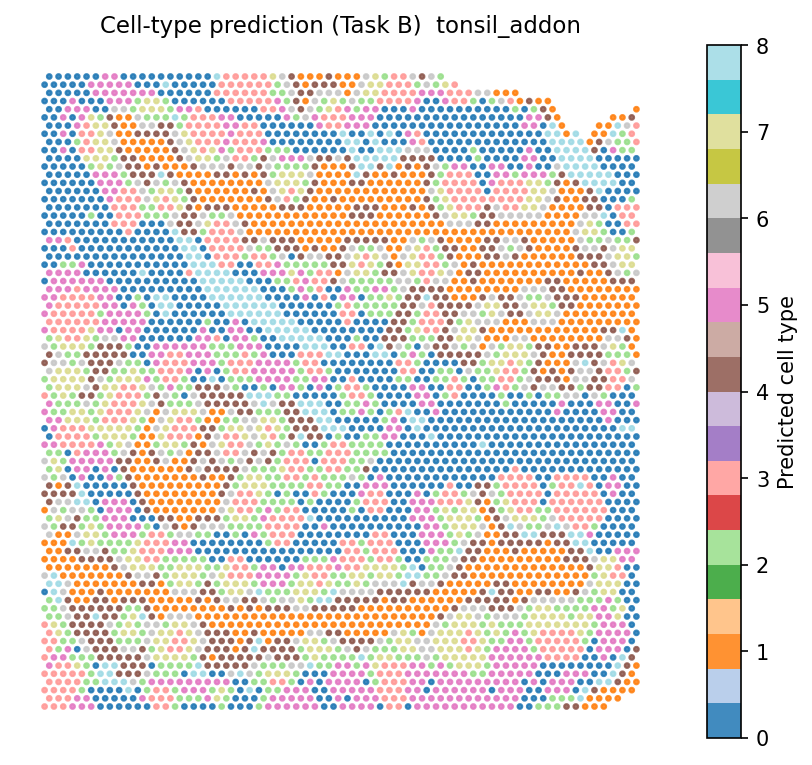}
    \caption{Human Tonsil Add-on}
\end{subfigure}
\caption{Task~B pseudo-cell-type prediction maps across four
CytAssist FFPE spatial proteomics datasets.}
\label{fig:taskb_maps}
\end{figure*}

Table~\ref{tab:taskb_summary} summarises the qualitative outputs of Task~B, and Fig.~\ref{fig:taskb_maps} shows the corresponding spatial pseudo-cell-type prediction maps. Because external ground-truth cell-type annotations are unavailable for these tri-modal datasets, we do not evaluate Task~B as a standalone supervised cell-type classification benchmark. Instead, the CellTypeHead serves as a pseudo-label-based auxiliary branch trained using Leiden-derived community labels. The resulting maps exhibit locally structured spatial patterns across all four datasets. Human Glioblastoma shows comparatively large and spatially coherent predicted regions, whereas Human Tonsil and Human Tonsil with Add-on Antibodies exhibit finer-grained and more intermixed patterns. These differences may reflect variation in tissue architecture and cellular heterogeneity across datasets. Overall, the maps provide qualitative evidence that the auxiliary Task~B objective encourages the shared latent representation to retain local pseudo-cell-type structure.

\begin{table*}[h]
\centering
\small
\caption{Task~B qualitative output summary.
Task~B is used as a semi-supervised regularisation branch rather than
a standalone cell-type benchmark because external ground-truth
cell-type annotations are not available.
The maps in Fig.~\ref{fig:taskb_maps} visualise the spatial
distribution of predicted pseudo-cell-type labels.}
\label{tab:taskb_summary}
\begin{tabular}{lcc}
\toprule
Dataset & Predicted classes & Spatial pattern \\
\midrule
Tonsil        & 11 & Mixed local domains \\
Breast Cancer & 15 & Region-level structure \\
Glioblastoma  & 11 & Spatially coherent regions \\
Tonsil Add-on &  9 & Fine-grained mixed labels \\
\bottomrule
\end{tabular}
\end{table*}

\subsection{Task C: Reasoning quality}
\begin{figure*}[t]
\centering
\includegraphics[width=\textwidth]{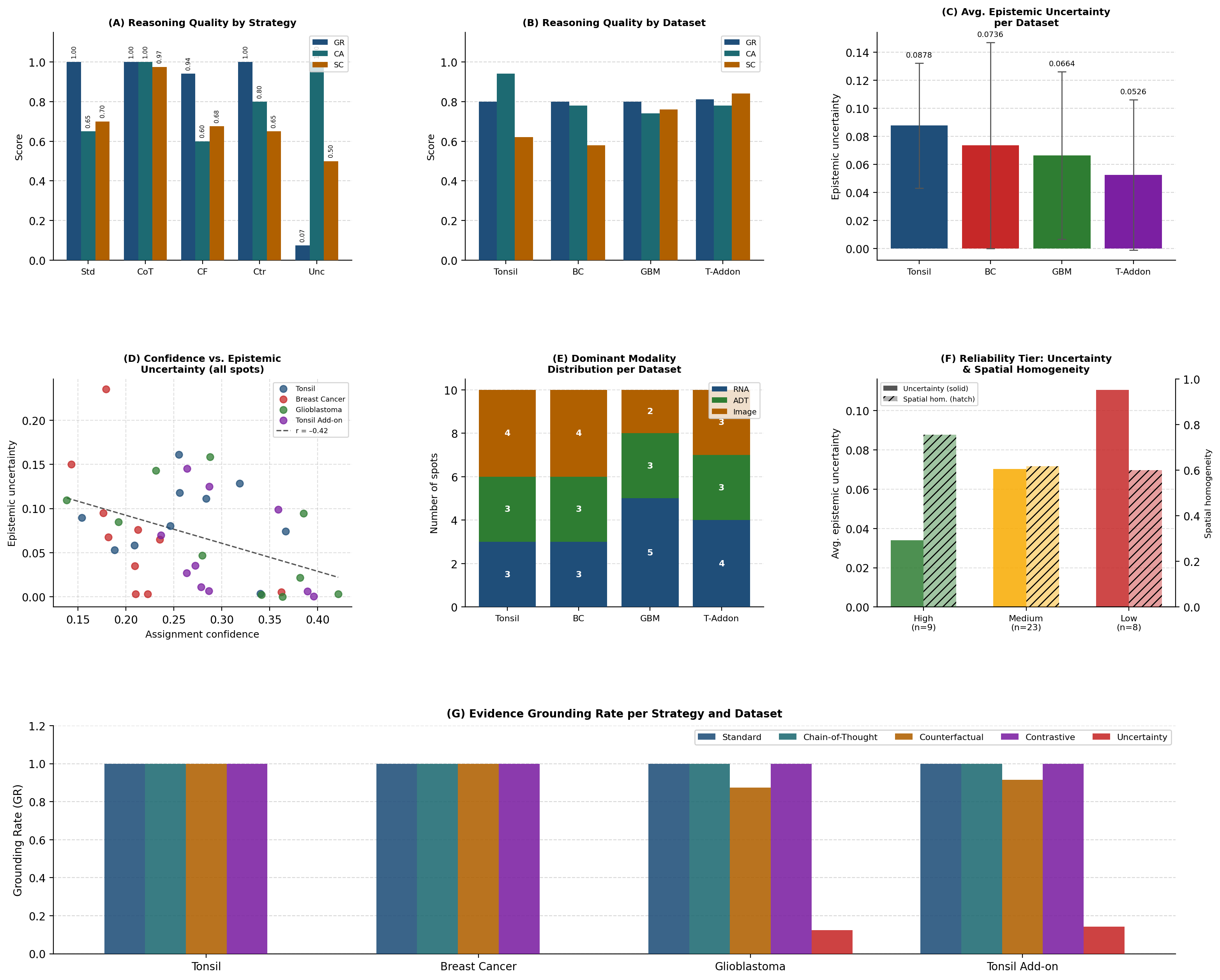}
\caption{Task~C reasoning analysis across four CytAssist FFPE datasets.
(A) Reasoning quality by strategy: Stepwise achieves the highest
overall faithfulness (GR = CA = 1.00, SC = 0.975); the uncertainty strategy
sacrifices gene grounding (GR = 0.074) to achieve perfect confidence
alignment (CA = 1.00).
(B) Reasoning quality by dataset.
(C) Average epistemic uncertainty per dataset; Tonsil Add-on achieves the
lowest uncertainty (0.053), consistent with its best clustering rank (1.22).
(D) Negative correlation ($r = -0.42$) between assignment confidence and
epistemic uncertainty across all 40 explained spots.
(E) Dominant modality distribution; Glioblastoma shows the highest RNA
dominance (5/10 spots), consistent with transcriptional heterogeneity at
tumour margins.
(F) Three-tier reliability audit: high-confidence spots exhibit $3.3\times$
lower epistemic uncertainty and higher spatial neighbourhood homogeneity
than low-confidence spots.
(G) Evidence grounding rate per strategy and dataset; the uncertainty
strategy consistently produces near-zero gene grounding by design.}
\label{fig:reasoning_analysis}
\end{figure*}

\paragraph{Strategy comparison.}
Table~\ref{tab:reasoning_strategy} and Figure~\ref{fig:reasoning_analysis}
report the three faithfulness metrics for the five reasoning strategies.
Each strategy is evaluated on the same 40 explained spots, comprising
10 spots from each of the four datasets. The Stepwise strategy achieves
the strongest joint performance (GR $=1.00$, CA $=1.00$, and SC $=0.975$).
Its structured five-stage format encourages the language model to
explicitly reference the supplied molecular, modality-routing, spatial,
and uncertainty evidence. 

The Standard and Contrastive strategies achieve full grounding rates, whereas the Counterfactual strategy also achieves a high grounding rate of 0.941. These results indicate that the evidence-only prompt design generally anchors the generated justifications to the supplied marker evidence. The Uncertainty strategy achieves a substantially lower GR of 0.074 because its prompt focuses primarily on routing uncertainty, confidence calibration, and assignment reliability rather than on marker-level evidence. GR is therefore less informative for this strategy. Nevertheless, the Uncertainty strategy achieves perfect confidence alignment (CA $=1.00$), consistent with its intended explanatory focus.

\begin{table}[t] 
\centering \small 
\caption{Reasoning quality by strategy. Each strategy is evaluated on the same 40 spots, comprising 10 spots from each dataset (200 generated justifications in total). GR = grounding rate; CA = confidence alignment; SC = spatial consistency.} 
\label{tab:reasoning_strategy} 
\begin{tabular}{lccc} 
\toprule Strategy & GR$\uparrow$ & CA$\uparrow$ & SC$\uparrow$ \\ 
\midrule Standard & 1.000 & 0.650 & 0.700 \\ 
Stepwise & \textbf{1.000} & \textbf{1.000} & \textbf{0.975} \\ 
Counterfactual & 0.941 & 0.600 & 0.675 \\ 
Contrastive & 1.000 & 0.800 & 0.650 \\ 
Uncertainty & 0.074 & \textbf{1.000} & 0.500 \\ 
\midrule Macro-average & 0.803 & 0.810 & 0.700 \\ 
\bottomrule 
\end{tabular} 
\end{table}

\paragraph{Dataset comparison.} 
Table~\ref{tab:reasoning_dataset} reports the reasoning metrics by dataset, aggregated across the five strategies. Human Tonsil with Add-on Antibodies achieves the highest spatial consistency (0.84) and the lowest average epistemic routing uncertainty (0.0526), indicating that its generated explanations are comparatively well aligned with the supplied neighbourhood evidence. Human Tonsil achieves the highest confidence alignment (0.94), indicating that its generated justifications most consistently match the assigned uncertainty tiers. These results reveal dataset-dependent differences in the grounding, calibration, and spatial support of the generated explanations.

\begin{table}[t] 
\centering \small 
\caption{Reasoning quality by dataset, aggregated across all five strategies. Ten spots are evaluated per dataset, corresponding to 50 generated justifications per dataset. Avg.~unc. = mean epistemic routing uncertainty per spot.} 
\label{tab:reasoning_dataset} 
\begin{tabular}{lcccc} 
\toprule 
Dataset & GR$\uparrow$ & CA$\uparrow$ & SC$\uparrow$ & Avg.~unc.$\downarrow$ \\ 
\midrule Tonsil & 0.800 & \textbf{0.940} & 0.620 & 0.0878 \\ 
Tonsil Add-on & \textbf{0.811} & 0.780 & \textbf{0.840} & \textbf{0.0526} \\ 
Breast Cancer & 0.800 & 0.780 & 0.580 & 0.0736 \\ 
Glioblastoma & 0.800 & 0.740 & 0.760 & 0.0664 \\ 
\bottomrule 
\end{tabular} 
\end{table}

\paragraph{Exploratory modality-routing patterns.} 
Among the 40 explained spots, RNA is the dominant routed modality for 7 of the 9 high-confidence spots, whereas image and ADT evidence occur equally often as the dominant modality among the 8 low-confidence spots. This pattern suggests that high-confidence assignments in the selected cases are more frequently supported by RNA evidence, whereas lower-confidence assignments more often depend on complementary histological or protein information. 

At the dataset level, RNA is the dominant modality for 50\% of the explained Glioblastoma spots. The tonsil datasets exhibit comparatively greater ADT dominance for spots whose supplied evidence includes B-cell-associated surface markers such as CD19, CD20, and IgD. Because the modality-routing weights are used as proxies rather than causal measures of modality contribution, these patterns should be interpreted as exploratory evidence of how OmicSync distributes information across modalities.
\subsection{The clustering-reasoning coupling}

\paragraph{Per-spot reliability audit.}
Table~\ref{tab:tiers} summarises the three-tier reliability analysis.
High-confidence spots ($\ge 0.35$; $n=9$) have a mean epistemic routing
uncertainty of 0.034, compared with 0.111 for low-confidence spots
($<0.20$; $n=8$). Thus, the low-confidence group exhibits approximately
$3.3\times$ the mean uncertainty of the high-confidence group.
High-confidence spots also show 26\% greater relative spatial neighbourhood
homogeneity than low-confidence spots (75.6\% versus 60.0\%). These patterns
are consistent with the numerical reliability signals reflecting meaningful
variation in assignment stability and local spatial support.

Assignment confidence and epistemic routing uncertainty exhibit a moderate
negative correlation ($r=-0.42$, $p<0.01$). The two measures are therefore
related but non-identical: confidence characterises the sharpness of the
cluster-assignment distribution, whereas routing uncertainty characterises the
variation of modality-selection weights under MC-Dropout sampling.

\begin{table}[t]
\centering
\small
\caption{Three-tier reliability audit across the 40 explained spots.
Low-confidence spots exhibit approximately $3.3\times$ the mean epistemic
routing uncertainty of high-confidence spots, whereas high-confidence spots
show greater spatial neighbourhood homogeneity.}
\label{tab:tiers}
\begin{tabular}{lcccc}
\toprule
Tier & Confidence & $n$ & Avg.~unc. & Spatial hom. \\
\midrule
High & $\ge 0.35$ & 9 & 0.034 & 75.6\% \\
Medium & $0.20$--$0.35$ & 23 & 0.070 & 61.7\% \\
Low & $< 0.20$ & 8 & 0.111 & 60.0\% \\
\bottomrule
\end{tabular}
\end{table}

\paragraph{Dataset-level patterns.}
Human Tonsil with Add-on Antibodies, which achieves the best average
clustering rank (1.22), also exhibits the lowest average epistemic routing
uncertainty (0.0526) and the highest reasoning spatial consistency (0.84).
This concordance suggests a relationship between stronger clustering
structure and more stable, spatially supported explanations on this dataset.

The relationship is not uniform across all datasets. For example, the separate
Human Tonsil dataset also achieves strong clustering performance, but has the
highest average routing uncertainty among the four datasets. Human Breast
Cancer exhibits the lowest spatial consistency (0.58), suggesting that its
explanations receive comparatively weaker support from local neighbourhood
composition, potentially because of heterogeneous tumour--stroma boundaries.
Overall, the coupled analysis provides a spot-level and dataset-level audit of
assignment confidence, modality routing, uncertainty, and spatial support,
rather than treating the generated justifications as unconstrained free-form
interpretations.

\subsection{Qualitative case studies}

Three case studies illustrate the coupling across the three modality regimes
observed in the data.

\paragraph{RNA-dominant: Glioblastoma Spot~3993 (Vascular,
conf.\ 42.1\%, unc.\ 0.003).}
RNA carries 98\% of the gate weight, and all five MC-Dropout samples agree,
yielding the lowest epistemic routing uncertainty in the glioblastoma cohort.
The top markers, C1QA and CAMK2N1, provide molecular support for the vascular
assignment, with C1QA reflecting complement-associated microenvironmental
signal and CAMK2N1 contributing additional vascular-associated evidence.
The homogeneous neighbourhood (5$\times$ Vascular) independently corroborates
the assignment. Counterfactual reasoning further indicates that removing either
the RNA modality or the C1QA/CAMK2N1 marker evidence would shift the prediction
toward Infiltrating margin (runner-up, 13.3\%), suggesting that both molecular
evidence and RNA availability are critical for this assignment.

\paragraph{Image-dominant: Tonsil Spot~420 (Vasculature,
conf.\ 34.0\%, unc.\ 0.0037).}
Image carries 96\% of the gate weight. Vessel-lumen morphology captured by the
UNI encoder dominates the transcriptomic context, whose neighbourhood
composition (2$\times$ GC, 1$\times$ FDC, 1$\times$ Naive B, 1$\times$ Plasma)
provides no clear vascular signal. The uncertainty-based explanation is
consistent with this interpretation: ``the low uncertainty (0.0037) indicates
confident modality selection; the model correctly prioritises histological
structure over local cellular context.''

\paragraph{ADT-dominant: Tonsil Add-on Spot~2078 (Macrophages,
conf.\ 39.6\%, unc.\ 0.0008).}
ADT carries 98\% of the gate weight, and the epistemic routing uncertainty is
0.0008, the lowest recorded across all 40 explained spots. The stepwise
reasoning attributes this assignment to the extended antibody panel:
``the richer protein signal enables unambiguous myeloid identification
that the standard tonsil panel cannot achieve at this certainty level.''
Direct comparison of the two tonsil datasets supports this interpretation:
the extended panel reduces average epistemic routing uncertainty from 0.088
in the standard tonsil dataset to 0.053 in the Add-on dataset, a
$1.7\times$ reduction. This quantifies the benefit of higher-plex protein
profiling for confident domain assignment.

\subsection{OmicSync-R: Reasoning-Guided Training}
\label{sec:results_r}

Table~\ref{tab:omicsync_r} reports the per-update reward trajectory of
OmicSync-R on Human Breast Cancer over 22 reasoning-update steps spanning
epochs 50--575. Table~\ref{tab:omicsync_r_clustering} reports the resulting
Task~A clustering metrics at $k=10$ compared with base OmicSync and GROVER.

\paragraph{Grounding rate.}
The evidence-grounding component of the REINFORCE reward is consistently
satisfied throughout training. Grounding rate equals 1.00 at 15 of the
22 update steps and falls below 0.93 only once (epoch 375, GR $=0.875$).
This indicates that the generated justifications remain strongly anchored to
the model-derived evidence across most reasoning updates. In particular, the
dominant modality and marker evidence used by the reasoning module remain
well aligned with the evidence exposed by the ClusteringHead, despite changes
in the clustering state over training.

\paragraph{Confidence alignment.}
Confidence alignment improves substantially from the initial update
(CA $=0.37$ at epoch 50) to a peak of CA $=0.775$ at epoch 225, with another
high value at epoch 350 (CA $=0.688$). This suggests that the ClusteringHead
progressively produces soft assignment distributions whose uncertainty is
better matched to the hedging language generated by the reasoning module.
The mid-training phase (epochs 175--375) achieves a mean CA of 0.646,
representing a $28.5\%$ improvement over the early phase
(epochs 50--150, mean CA $=0.503$). A transient degradation is observed at
epoch 425 (CA $=0.313$), after which confidence alignment recovers to
CA $\approx0.51$--$0.54$ through the end of training.

\paragraph{Spatial consistency.}
Spatial consistency is the most variable component of the reward. The highest
values (SC $=0.50$) are observed at epochs 175 and 350, coinciding with high
reward episodes. Spatial consistency declines in the late training phase
(mean SC $=0.21$ over epochs 400--575), suggesting that neighbourhood
composition becomes less predictive of the final assignments as the soft
assignments continue to sharpen. This behaviour is consistent with
late-stage DEC-style clustering, where probability mass concentrates on
individual centroids and local neighbourhood composition may exert less direct
influence on the assignment decision.

\paragraph{Overall reward trajectory.}
The exponential moving-average (EMA) reward baseline, used to reduce
REINFORCE gradient variance, rises from 0.505 at epoch 50 to a peak of
0.585 at epoch 375 before stabilising near 0.550 through the end of training.
Peak mean reward is attained at epoch 225 ($\bar{R}=0.706$, GR $=1.00$,
CA $=0.775$, SC $=0.417$), followed closely by epoch 350
($\bar{R}=0.700$, GR $=1.00$, CA $=0.688$, SC $=0.500$). Across all
22 updates, the mean reward is $0.573\pm0.075$, remaining above the
initial baseline of 0.50 on average. These results indicate that the
REINFORCE signal provides a usable auxiliary learning signal for
reasoning-guided training, while the clustering metrics in
Table~\ref{tab:omicsync_r_clustering} quantify its downstream effect on
Task~A performance.

\paragraph{Impact on Task~A clustering metrics.}
Table~\ref{tab:omicsync_r_clustering} shows that OmicSync-R improves over
base OmicSync on eight of nine metrics at $k=10$: ARI increases from 45.73
to 46.72 ($+0.99$), FMI from 51.75 to 53.54 ($+1.79$), AMI from 43.72 to
45.79 ($+2.07$), Jaccard from 35.70 to 38.78 ($+3.08$), SilC from 28.92 to
30.20 ($+1.28$), Purity from 56.78 to 58.13 ($+1.35$), CHI from 2990 to
2995, and DBI from 178.76 to 165.30 ($-13.46$, lower is better). NMI
declines modestly from 54.10 to 52.20 ($-1.90$). Using the GROVER results
reported for Human Breast Cancer in the main baseline comparison
(Table~\ref{tab:clustering_results}), OmicSync-R also surpasses GROVER on
ARI, CHI, AMI, Jaccard, SilC, and Purity. These results suggest that the
reasoning-quality reward provides a useful auxiliary learning signal on
Human Breast Cancer, improving most external agreement metrics and several
internal clustering metrics at the selected value $k=10$.

\paragraph{$k$-sensitivity.}
The benefit of OmicSync-R is concentrated at $k=10$. For $k=6$--$9$,
OmicSync-R yields ARI values of 23.68--27.77, substantially below its ARI
of 46.72 at $k=10$. This indicates that the reasoning-guided update is
sensitive to the number of cluster centroids. The interaction between the
REINFORCE reward and centroid selection therefore warrants further
investigation.

\paragraph{Interpretation.}
The results show that OmicSync-R maintains high grounding fidelity
(GR $\approx 0.98$ on average) and can substantially improve confidence
alignment during training, increasing from CA $=0.37$ at the first update to
a peak of CA $=0.775$. By contrast, spatial consistency remains unstable,
suggesting that neighbourhood-level coherence is harder to optimise with a
reward signal computed on individual spot samples. Improving this component
may require a graph-level or neighbourhood-aware reward formulation. Overall,
these findings support OmicSync-R as a proof-of-concept showing that
reasoning-quality scores can be used as a non-differentiable auxiliary
training signal for spatial domain clustering.

\begin{table}[h]
\centering
\small
\renewcommand{\arraystretch}{1.15}
\caption{OmicSync-R reward trajectory on Human Breast Cancer
(selected epochs).
GR = grounding rate; CA = confidence alignment;
SC = spatial consistency; $\bar{R}$ = mean composite reward;
$b_t$ = EMA baseline.
All 22 update steps are reported in the supplementary material.}
\label{tab:omicsync_r}
\begin{tabular}{rccccr}
\toprule
Epoch & GR$\uparrow$ & CA$\uparrow$ & SC$\uparrow$
      & $\bar{R}\uparrow$ & $b_t$ \\
\midrule
 50  & 0.980 & 0.370 & 0.450 & 0.551 & 0.505 \\
100  & 1.000 & 0.578 & 0.481 & 0.650 & 0.517 \\
175  & 1.000 & 0.589 & 0.500 & 0.661 & 0.541 \\
\textbf{225} & \textbf{1.000} & \textbf{0.775} & 0.417
             & \textbf{0.706} & 0.555 \\
\textbf{350} & \textbf{1.000} & 0.688 & \textbf{0.500}
             & \textbf{0.700} & 0.581 \\
425  & 0.925 & 0.313 & 0.229 & 0.437 & 0.571 \\
575  & 1.000 & 0.538 & 0.250 & 0.553 & 0.550 \\
\midrule
\multicolumn{4}{l}{\textit{Mean} (all 22 updates)}
      & 0.573 & --- \\
\multicolumn{4}{l}{\textit{Std.\ dev.}}
      & 0.075 & --- \\
\bottomrule
\end{tabular}
\end{table}

\begin{table}[h]
\centering
\small
\setlength{\tabcolsep}{6pt}
\renewcommand{\arraystretch}{1.15}
\caption{OmicSync-R clustering results on Human Breast Cancer
($k=10$) compared to base OmicSync.
$\Delta$ = OmicSync-R minus OmicSync.
\textbf{Bold} = best.}
\label{tab:omicsync_r_clustering}
\begin{tabular}{lrrr}
\toprule
Metric & OmicSync & OmicSync-R & $\Delta$ \\
\midrule
ARI$\uparrow$     & 45.73 & \textbf{46.72} & $+0.99$ \\
NMI$\uparrow$     & \textbf{54.10} & 52.20         & $-1.90$ \\
FMI$\uparrow$     & 51.75 & \textbf{53.54} & $+1.79$ \\
SilC$\uparrow$      & 28.92 & \textbf{30.20} & $+1.28$ \\
AMI$\uparrow$     & 43.72 & \textbf{45.79} & $+2.07$ \\
Jaccard$\uparrow$ & 35.70 & \textbf{38.78} & $+3.08$ \\
CHI$\uparrow$     & 2990  & \textbf{2995}  & $+5$ \\
Purity$\uparrow$  & 56.78 & \textbf{58.13} & $+1.35$ \\
DBI$\downarrow$   & 178.76 & \textbf{165.30} & $-13.46$ \\
\bottomrule
\end{tabular}
\end{table}

\section{Discussion}
\label{sec:discussion}

\paragraph{Coupled audit trail.}
The novelty of OmicSync lies not in any individual head alone, but in the
\emph{coupled audit trail}: the clustering pipeline exposes interpretable
signals, including assignment confidence, modality-routing weights, marker
evidence, uncertainty estimates, and neighbourhood composition. These signals
are then used by the reasoning module to produce human-understandable,
evidence-grounded explanations of each spatial-domain assignment.

\paragraph{Reasoning as a training signal.}
OmicSync-R provides a proof-of-concept that reasoning-quality scores can serve
as a non-differentiable auxiliary training signal through REINFORCE. On Human
Breast Cancer, the most challenging dataset, OmicSync-R improves ARI from
45.73 to 46.72 and improves eight of nine clustering metrics at $k=10$.
These results suggest that evidence-grounded reasoning coherence and spatial
domain quality can be partially aligned during training, rather than treated
as entirely separate post-hoc objectives. Because the iterative REINFORCE loop
requires repeated calls to the reasoning module, OmicSync-R is more
resource-intensive than post-hoc explanation; we therefore restrict the current
evaluation to one dataset. Thus, the OmicSync-R experiment should be viewed as
an initial demonstration rather than a complete validation.

\paragraph{Adaptive spatial smoothing.}
The adaptive $n_\mathrm{hops}$ mechanism addresses the ARI/SilC trade-off in
topology-aware contrastive learning: homogeneous tissue can tolerate larger
exclusion radii, improving cluster separation, whereas heterogeneous tissue
requires smaller radii to preserve fine-grained spatial boundaries. This
reduces the need for a manually fixed smoothing configuration across datasets
with different spatial organization.

\paragraph{Limitations.}
Base OmicSync's Task~C operates after model training and does not
back-propagate into the clustering objective; it explains rather than
optimises the partition. OmicSync-R addresses this limitation through
REINFORCE, but the reward is limited to three automatically computable
faithfulness metrics, exhibits sensitivity to the number of clusters $k$, and
requires repeated reasoning-module calls during training. This iterative
self-improvement loop can be resource intensive. Task~B pseudo-labels are
RNA-derived, so the reported analysis evaluates the regularisation benefit of the
auxiliary classification task rather than external classification accuracy.
The reasoning module is constrained by the evidence supplied by the model, so
its explanations are limited to the five supplied evidence types, namely, 
assignment confidence, modality-routing
weights, marker evidence, uncertainty estimates, and neighbourhood composition,
and inherit any upstream uncertainty in those signals.

\paragraph{Future work.}
Future work will extend OmicSync-R to all four datasets, investigate
graph-level or neighbourhood-aware reward formulations to improve spatial
consistency, and explore reward annealing or multi-$k$ training to reduce
$k$-sensitivity. Extending the framework to additional modalities, such as
ATAC-seq chromatin accessibility and spatial metabolomics, and to
higher-resolution spatial platforms, such as Xenium and MERFISH, is also a
natural direction. Future work will also explore amortised reward models, cached reasoning
evaluations, and smaller verifier models to reduce the computational cost of
iterative reasoning-guided training.

\section{Conclusion}
\label{sec:conclusion}

We presented OmicSync, a spatial multi-omics framework that couples
domain clustering with five-strategy LLM reasoning through a shared
set of per-spot interpretability signals, and OmicSync-R, a
reasoning-guided training variant that uses REINFORCE with
automatically computed reasoning-quality scores as reward signals.

OmicSync achieves the best average rank on three of four CytAssist
FFPE benchmarks and provides assignment confidence, epistemic
uncertainty, modality attribution, and auditable natural-language
justification for spatial-domain assignments, capabilities not jointly
provided by competing methods. 
OmicSync-R further improves ARI on Human Breast Cancer, the only benchmark
where base OmicSync does not achieve the best overall rank, from 45.73 to
46.72 and surpasses GROVER, the most competitive non-OmicSync baseline, on
six of nine metrics, suggesting that evidence-grounded reasoning coherence
and clustering quality can be partially aligned during training.
Together, these contributions advance spatial omics analysis from
opaque partitioning toward principled, interpretable, and
reasoning-guided tissue domain discovery.

\bibliographystyle{unsrtnat}
\bibliography{omicsync}

\end{document}